\documentclass[10pt]{article}
\usepackage[preprint]{tmlr}

\usepackage{microtype}
\usepackage{graphicx}
\usepackage{amsmath,amssymb,amsthm}
\usepackage{booktabs}
\usepackage{array}
\usepackage{xcolor}
\usepackage{hyperref}
\usepackage{url}
\makeatletter\g@addto@macro{\UrlBreaks}{\do\-}\makeatother
\hypersetup{hidelinks,
  pdftitle={Richness Is Not Evenness: Ordered Renyi Crossings under Self-Training},
  pdfauthor={Pranjal Srivastava}}

\graphicspath{{figures/}}

\theoremstyle{plain}

\newtheorem{lemma}{Lemma}
\newtheorem{proposition}{Proposition}

\theoremstyle{definition}

\newcommand{\V}{\mathcal{V}}
\newcommand{\E}{\mathbb{E}}
\newcommand{\Var}{\operatorname{Var}}
\newcommand{\Cov}{\operatorname{Cov}}
\newcommand{\KL}{\operatorname{KL}}
\newcommand{\ph}{\hat p}
\newcommand{\Cbar}{\bar C}
\newcommand{\pbar}{\bar p}
\newcommand{\Htwo}{H_2}
\newcommand{\Hz}{H_0}
\newcommand{\Hinf}{H_\infty}

\title{Richness Is Not Evenness:\\Ordered R\'enyi Crossings under Self-Training}

\author{\name Pranjal Srivastava \\
 \addr University of Michigan}

\def\month{10}
\def\year{2026}

\begin{document}
\maketitle
\begingroup
\renewcommand\thefootnote{}
\footnotetext{\textit{AI assistance:} Claude Code and OpenAI Codex (details before references);
the author is responsible for all claims. \textit{Code/data:} \href{https://github.com/pranjal-srivas/CollisionEntropySelfTraining}{github.com/pranjal-srivas/CollisionEntropySelfTraining}
(currently private; public release planned after arXiv announcement).}
\addtocounter{footnote}{-1}
\endgroup

\begin{abstract}
Recursive self-training of language models, a setting studied in model collapse, can reduce
vocabulary while making frequent words, mostly function words, less dominant.
We fine-tune English GPT-2 on its own output for nine generations, vary the training-data
sampling temperature, and measure every model with one fixed sampler. As training temperature
rises, R\'enyi entropy changes turn from loss to gain at distinct temperatures: min-entropy
first, then collision entropy, Shannon entropy and observed richness. Between the outer
crossings, word types fall while the most-common word becomes less dominant. Untruncated
training at $T=1$ lies in this interval: at 13,000 words, generation~9 has 24\% fewer types than
generation~0 yet collision entropy is 0.91 bits higher. The ordering persists with a finer grid,
different measurement samplers, a fixed word budget and five fresh training seeds. Min-entropy again crosses
before richness in four further settings (in Hindi only at interpolated roots) and in 12 of
13 distinct measurements under eight text representations. Next-token identities and a
categorical simulation isolate decoding and sampling effects; the neural crossing order
remains empirical. Fine-tuning on human text can produce the same split while likelihood
improves, and no single order's sign reliably identifies worse likelihood across the tested
settings. We recommend reporting entropy profiles at stated budgets with an independent
quality measure.
\end{abstract}

\section{Introduction}
\label{sec:intro}

Training language models on synthetic text can erode quality and diversity
\citep{shumailov2024,dohmatob2024}. Single diversity scores measure different properties:
richness counts observed word types, whereas higher-order entropies increasingly emphasize
common words. Losing rare words while spreading usage more evenly among those retained can
therefore lower richness and raise higher-order entropy. The reported direction depends on
which property is measured.

\begin{figure}[t]
\centering
\includegraphics[width=\linewidth]{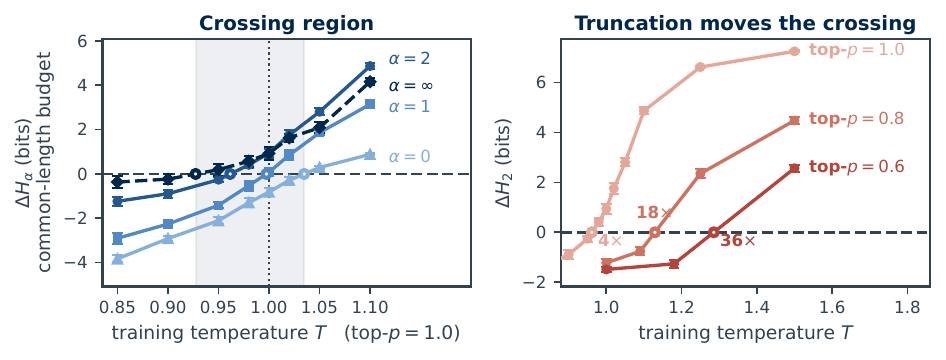}
\caption{English GPT-2 after nine rounds of recursive full fine-tuning, five seeds per cell.
Changes compare generation~9 with generation~0 under the fixed measurement sampler ($T=1$,
top-$p=0.95$) at common-length budgets. \emph{Left:} measured means and seed standard
deviations near the crossing region. Open circles mark linearly interpolated roots; inside the
shaded interval richness falls while min-entropy rises, as measured at four temperatures.
\emph{Right:} collision-entropy changes at three training top-$p$ values. Labels give the
cross-perplexity ratio interpolated at each root; no model was trained at exactly those
temperatures.}
\label{fig:crossings}
\end{figure}

We fine-tune English GPT-2 on its own samples for nine generations, varying the temperature
and nucleus (top-$p$) threshold of the \emph{training sampler}. Every model is measured with
one fixed sampler ($T=1$, top-$p=0.95$). As training temperature rises, min-entropy $\Hinf$
turns from loss to gain first, near $T=0.927$, followed by collision entropy $\Htwo$
($0.962$), Shannon entropy $H_1$ ($0.997$) and observed richness $\Hz$ ($1.035$)
(Figure~\ref{fig:crossings}). Between the outer crossings, word types fall while the most-common
word becomes less dominant. Untruncated sampling at $T=1$ lies inside this interval.

The outer ordering survives 12 of 13 distinct measurements under eight representations and
recurs under LoRA, in OLMo and in German and Hindi GPT-2 (Hindi only by interpolation).
We test the diagnostic value of these changes: truncation lowers entropy relative to untruncated
training in ten settings, but no order's sign reliably identifies worse likelihood, and human
adaptation can produce the same split while likelihood improves.

Diversity profiles are standard \citep{hill1973,tothmeresz1995}.
\citet{drayson2025detection} already observed rising $n$-gram diversity with worsening perplexity
in recursive GPT-2 under pure sampling ($T=1$, no truncation). Our contribution is the measured
ordering of sign changes across entropy orders, its sensitivity to measurement choices, and
tests of its diagnostic limits. Five fresh training seeds replicate the primary ordering
(Appendix~\ref{app:freshseeds}). Section~\ref{sec:crossings} presents the main crossing
result; the later experiments test its transfer and diagnostic limits, with detailed controls
in the appendices.

\section{Background and related work}
\label{sec:related}

\paragraph{Diversity profiles.} R\'enyi entropy \citep{renyi1961}, extended by limits at orders zero
and one, includes observed log-richness, Shannon entropy \citep{shannon1948} and collision entropy. Yule's $K$ has been studied as a constancy measure of text \citep{tweedie1998}. Collision
entropy is closely related and has also been studied in this role \citep{tanakaishii2015}. In ecology, $\sum_i p_i^2$ is
Simpson's concentration \citep{simpson1949} and $2^{H_\alpha}$ are Hill numbers
\citep{hill1973,jost2006}; reporting a profile over orders, whose members weight rare and abundant species differently, is
long-standing practice \citep{tothmeresz1995}. The Vendi Score and its order-indexed
extensions apply the same idea to the spectra of similarity matrices
\citep{friedman2023vendi,pasarkar2024cousins}, coverage-rarefied Hill--Shannon diversity has been
used to compare the claims made by language models with web search \citep{wright2026epistemic},
and higher-order R\'enyi entropies of repeated sequences separate model text from natural language
\citep{tanakaishii2026repeated}. \citet{zouhar2023}
sweep R\'enyi orders to evaluate tokenizers, and \citet{france2026} find that entropy- and
concentration-based measures can rank human and model text differently. Recommending a profile
is therefore not new; our question is how the \emph{sign} of each order's change behaves under
controlled recursion.

\paragraph{Generation and recursive training.} Distinct-$n$ \citep{li2016} and self-BLEU
\citep{zhu2018} measure output diversity, MAUVE \citep{pillutla2021} measures distributional
fit to reference text, and diversity metrics differ in whether they capture form or content
\citep{tevet2021diversity}. Recursive degradation has been studied empirically and through
scaling laws \citep{shumailov2024,dohmatob2024}, also in image models \citep{alemohammad2023},
including effects on
lexical and syntactic diversity \citep{guo2024} and repetition \citep{herel2024collapse}; even
without recursion, synthetic training data over-concentrate $n$-gram features \citep{zhu2025synthesize}. Curation, the share of fresh data, and
whether synthetic data replace or accumulate with real data all change the outcome
\citep{ferbach2024curated,briesch2023,gerstgrasser2024collapse,kazdan2025collapse}.
Next-token concentration has been used to track collapse, with perplexity-based filtering as a
remedy \citep{gambetta2026surprise}, and the time a model takes to forget a symbol has been
analyzed theoretically \citep{suresh2025forgetting}. Definitions of collapse vary
\citep{schaeffer2025position}. We define degradation by a stated reference-likelihood cutoff and
evaluate diversity changes separately.

\paragraph{Closest prior results.} \citet{seddik2024howbad} derive the collision ratchet, with
its rate, for recursive maximum-likelihood training of categorical models;
Proposition~\ref{prop:ratchet} (Appendix~\ref{app:proofs}) extends the argument to upper-set
truncation. \citet{dohmatob2024} model truncation as cutting the tail of the distribution and
low temperature as narrowing it, and \citet{marchi2024heat} prove that closed-loop retraining at a
non-trivial sampling temperature degenerates, either to a few outputs or to a uniform ``heat
death'', unless enough fresh data enter each round. Under untruncated, infinite-sample recursive tempering of a
non-uniform categorical law, all positive orders switch sign at $T=1$ and support size stays
fixed (Propositions~\ref{prop:temper} and~\ref{prop:sens}). The separate neural roots depart
from this baseline, without isolating the effects of sampling, autoregressive contexts,
refitting or the fixed truncated reader.
\citet{drayson2025detection} vary the decoder
in recursive GPT-2: under pure sampling ($T=1$, no truncation), $n$-gram diversity rises while
perplexity worsens; all five other decoders lose $n$-gram diversity. Meanwhile,
\citet{kovac2025} link lexical diversity to recursive distribution shift. We measure the order
of zero crossings across the R\'enyi spectrum, test its
sensitivity to budget and representation, and ask what it lets a diagnostic infer.

\paragraph{Recent and concurrent work.} \citet{hu2026profiles} independently recommend
diversity profiles across orders and scales. \citet{skobelev2026finetuning} measure diversity by collision
probability, bound the absolute model--target collision gap by the square root of the KL
divergence from target to model, and show that supervised fine-tuning can leave a model
under- or over-dispersed. Their setting is fine-tuning toward a fixed target at order~2;
ours tracks the sign changes across R\'enyi orders during recursive self-training. Other studies address diversity loss in
post-training \citep{karouzos2026where}, entropy-based early warnings \citep{zhang2026silent},
recursion in larger models \citep{nakamura2026demystifying}, the loss of rare forms within
languages \citep{vanmassenhove2026tail}, ecological diversity measures for linguistic structure
\citep{gude2026aligned}, and the diversity of synthetic data sources
\citep{schaffelder2026eggs}. Concurrently, \citet{liu2026fragility} trace rapid lexical collapse
in vLLM to draws shared across requests and replayed across generations. Changing seeds removes
the rapid loss in their controls, revising their earlier account. Our Transformers harness
keys streams by replicate, generation and phase, sharing them across paired arms but not
replaying an unchanged seed across generations. \citet{liu2026restructures} finds that self-training amplifies
surface markers while deeper syntactic structures decay, and that 100-word-window type--token
ratio rises. In our own English GPT-2 runs at $T=1$, the seed-mean ratio over non-overlapping
100-word windows at common length rises from $0.746$ to $0.841$ while the common-length type inventory falls by
43\%.

\section{The measure}
\label{sec:measure}

\paragraph{Setup.} Let $p=(p_i)_{i\in\V}$ be a distribution over a vocabulary, either a
next-token softmax or the empirical word frequencies $\ph_v=f_v/N$ of a text. For finite
$\alpha\ge0$ with $\alpha\ne1$, the R\'enyi entropy is
$H_\alpha(p)=\frac{1}{1-\alpha}\log_2\sum_{i:p_i>0}p_i^\alpha$. Thus
$H_0=\log_2|\operatorname{supp}p|$, and the limiting orders are Shannon entropy
$H_1=-\sum_i p_i\log_2p_i$ and min-entropy $H_\infty=-\log_2\max_i p_i$. Entropy does not
increase with the order, so $H_0\ge H_1\ge\Htwo\ge H_\infty$ (Appendix~\ref{app:proofs}).
Collision entropy is $\Htwo=-\log_2 C$, where $C(p)=\sum_i p_i^2$ is the probability that two
independent draws coincide.

\paragraph{Yule's $K$.} In empirical-probability form \citep{yule1944},
\begin{equation}
K = 10^4\,\frac{\sum_v f_v^2 - N}{N^2}
 = 10^4\Big(2^{-\Htwo(\ph)}-\tfrac1N\Big).
\label{eq:yulek}
\end{equation}
By \eqref{eq:yulek}, at fixed $N$ Yule's $K$ is a strictly decreasing function of $\Htwo$, so
for texts of equal length a fall in $K$ and a rise in $\Htwo$ are the same event. The $1/N$
offset can break this correspondence between texts of different lengths, which is one reason we
compare texts at matched budgets. The offset $10^4/N$ vanishes as $N$ grows, so for long texts
$K\approx10^4\cdot2^{-\Htwo(\ph)}$.

\paragraph{Measurement.} For a token string $x_{1:N}$ we compute $\Htwo(x_{1:N})=\Htwo(\ph)$
from its observed counts, so no model probabilities or logits are needed. Higher orders
emphasize common types, whereas $\Hz$ only counts observed types. We use ``evenness'' loosely
for this sensitivity to concentration; $\Htwo$ also depends on richness and is not normalized
by the size of the support.

Order~2 has Simpson's unbiased pair-count estimator \citep{simpson1949}. Under i.i.d.\ sampling,
$\widehat C_{\mathrm u}=\sum_v f_v(f_v-1)/[N(N-1)]$ is unbiased for the collision probability,
and $K=10^4(N-1)\widehat C_{\mathrm u}/N$. This does not make $-\log_2\widehat C_{\mathrm u}$
an unbiased estimate of entropy. Integer orders above two have analogous falling-factorial
estimators, fixed-sample counterparts of those analyzed under Poisson sampling by
\citet{acharya2017renyi}
(Appendix~\ref{app:factorial-estimator}).

\section{Effects of decoding on next-token entropy}
\label{sec:pertoken}

For standard temperature and nucleus controls \citep{holtzman2020}, the following facts
are known. Majorization orders tempered distributions \citep{marshall2011inequalities},
implying temperature monotonicity of positive-order R\'enyi entropies
\citep{mattei2026welltempered}. \citet{nadeem2020systematic} give the Shannon entropy-reduction
counterparts for top-$k$, nucleus and low-temperature sampling;
\citet[App.~J]{skobelev2026finetuning} concurrently state order-2 temperature monotonicity
and a support-size cap. We restate the order-2 facts, with proofs in Appendix~\ref{app:proofs},
for the document-level analysis in \S\ref{sec:vbridge}.

\begin{proposition}[Temperature monotonicity]
\label{thm:temp}
For a softmax $p_i(T)\propto e^{z_i/T}$ over finite, non-constant logits, $\Htwo(p_T)$ is strictly increasing in $T$ on $(0,\infty)$, with $\Htwo\to\log_2 m$ as $T\to 0^+$, where $m$ is the number of tied maxima (so the limit is $0$ if the maximum is unique), and $\Htwo\to\log_2|\V|$ as $T\to\infty$.
\end{proposition}
With $\beta=1/T$, $I\sim p_T$ and $\mu=\E[z_I]$,
$\frac{d}{d\beta}\sum_i p_i^2 = 2\sum_i p_i^2(z_i-\mu)=2\Cov(p_I,z_I)>0$, because $p_i$
increases with $z_i$. Raising the temperature therefore lowers the next-token collision
probability.

\begin{proposition}[Truncation strictly lowers $\Htwo$]
\label{thm:trunc}
Let $S$ be an upper set, meaning $\min_{i\in S}p_i\ge\max_{j\notin S}p_j$ (top-$k$ and top-$p$ selection both produce one), with retained mass $a=\sum_{i\in S}p_i\in(0,1)$, and let $q_i=p_i/a$ on $S$. Then $\Htwo(q)<\Htwo(p)$.
\end{proposition}
Conditioning on an upper set gives $\E[p_I\mid I\in S]\ge\E[p_I]$ (Lemma~\ref{lem:upperset}),
and renormalizing then yields $C(q)=a^{-1}\E[p_I\mid I\in S]>C(p)$. Keeping the top 20 of 100
equally likely tokens already costs $\log_2 5\approx2.3$ bits.

\begin{proposition}[Fixed-support cap]
\label{prop:cap}
If sampling is restricted to a set $S$ with $|S|=k$, then for any reshaping of the surviving mass (for example, any temperature applied within $S$), $\Htwo\le\log_2 k$, with equality if and only if the distribution is uniform on $S$.
\end{proposition}
The cap applies to a \emph{fixed} retained set. Raising the temperature before top-$p$
selection can enlarge that set, and the retained set differs from context to context, so the
cap by itself does not bound document-level entropy independently of temperature.

\section{From softmax to document: the \texorpdfstring{$V$}{V}-bridge}
\label{sec:vbridge}

Let $p_{c_t}$ be the conditional token distributions at $N$ contexts and
$\pbar=N^{-1}\sum_t p_{c_t}$ their mean. We first relate the collision entropy of this mean to
the variation across contexts, and then connect it to sampled frequencies.

\begin{lemma}[Variance decomposition]
\label{lem:vbridge}
With mean per-token collision $\Cbar=\frac1N\sum_t\sum_v p_{c_t}(v)^2$ and across-context dispersion $V=\sum_v\Var_t[p_{c_t}(v)]\ge0$, where $\Var_t$ averages over the $N$ contexts,
\begin{equation}
C(\pbar)=\Cbar - V,\qquad \Htwo(\pbar)=-\log_2(\Cbar-V).
\label{eq:vbridge}
\end{equation}
\end{lemma}

This is the additive Gini--Simpson partition \citep{rao1982,lande1996} applied to conditional
token distributions: it separates within-context concentration from between-context variation. Deterministic conditionals have zero per-token entropy, yet their mean
can spread its mass over many different argmaxes. It also explains why the next-token results
of \S\ref{sec:pertoken} do not fix the direction of document-level change: a change of
temperature changes both $\Cbar$ and $V$, and $\Htwo(\pbar)$ rises exactly when
$\Cbar-V$ falls.

For a token process of fixed length $N$, let $p_{c_t}$ be the \emph{sampling law} given the
history and $\ph_N$ the empirical frequencies. Each one-hot sampled token minus its sampling law is
a martingale difference, so $\E\|\ph_N-\pbar_N\|_2^2\le 1/N$ (Appendix~\ref{app:proofs}). The bound compares empirical
frequencies with the average sampling law along the same random path; fixed-population
convergence and autoregressive entropy bias require separate arguments.

\paragraph{Fixed-context measurement.} On softmax probabilities of Qwen2.5-3B-Instruct at 200
fixed 48-token contexts from four Gutenberg books (Appendix~\ref{app:figvals}), $T=0.1$ gives
$\Cbar=0.90$ and $V=0.88$, so $\Htwo(\pbar)=5.65$ bits, far above the value $-\log_2\Cbar=0.14$
bits implied by the mean per-context collision; computing $V$ directly reproduces \eqref{eq:vbridge} to within $1.1\times10^{-8}$ at every temperature.

\section{Experiments}
\label{sec:experiments}

Word-level R\'enyi values apply NLTK \texttt{word\_tokenize} \citep{bird2009nltk}, version 3.9.1,
to the original text, lowercase the resulting tokens, and drop tokens that contain no letter or
number; punctuation attached to a
word stays with it. The same English-trained tokenizer is applied to German, Spanish and Hindi.
Chinese uses Han characters
(U+4E00--U+9FFF). Next-token softmax values use each model's byte-pair encoding (BPE) vocabulary. We
compare changes only within one representation.

\subsection{Setup}
\label{sec:setup}

\paragraph{Protocol.} Each generation samples a training corpus from the current model and
fine-tunes it on that corpus. We vary training temperature $T$ and top-$p$; measurements use
one fixed \emph{measurement sampler} ($T=1$, top-$p=0.95$, top-$k$ disabled; the ``probe''
in the code), holding the reader constant across trained models.
We call top-$p=0.80$ at $T=1$ the \emph{truncated} arm and top-$p=1$ at $T=1$ the
\emph{untruncated} arm. Random numbers are shared across arms: each draw is keyed by seed,
generation and phase, and all arms use the same pool of real-text prompts. The LoRA \citep{hu2022lora}
arms of a replicate start from one saved initial adapter, verified by a SHA-256
fingerprint of its trainable tensors. PEFT's zero-initialized $B$ matrices
\citep{mangrulkar2022peft} yield identical initial outputs even with different random $A$ matrices.

\paragraph{Budgets and seeds.} Richness depends strongly on the length of a text, so every
comparison states its budget. A \emph{common-length} comparison truncates the generation-0 and
generation-9 texts of each seed to their shared length, and a \emph{fixed-budget} comparison
uses the first 13,000 words of every text. Most studies use three seeds; the primary English grid and its refinements, the OLMo and Hindi grids, the
Pythia study and the checkpoint-branch intervention use five (Pythia's first stage was later extended to
eight, and the independent-rerun comparison in Appendix~\ref{app:detect} used ten). Held-out cross-perplexity is the exponential of mean next-token negative log
likelihood on real text disjoint from training.

\paragraph{Boundaries and sampling defaults.} Before training samples are concatenated, the
harness removes the padding-token ID. For GPT-2 this also removes the end-of-sequence (EOS)
token. OLMo-2 retains its distinct EOS, and Qwen2.5-Instruct retains its primary EOS but removes
its second stop ID, which equals padding. The German and Hindi checkpoints declare an ordinary subword as their EOS,
so their continuations end only when that subword happens to be sampled
(Appendix~\ref{app:repr}). Every sampling call in the training harness passes the temperature,
top-$p$ and top-$k=0$ explicitly and otherwise inherits the checkpoint's generation configuration. For
Qwen2.5-0.5B-Instruct that configuration sets a repetition penalty of 1.1, which therefore
applied to every Qwen-Instruct training corpus and measurement. No other sampled checkpoint sets one;
\S\ref{sec:richness} tests its effect with a matched rerun.

\paragraph{Models and data.} The primary model under full fine-tuning (full-FT) is GPT-2 with 124M parameters (listed as 117M
in \citealt{radford2019}). Robustness settings use German \citep{dbmdz2020germangpt2} and Spanish
\citep{flaxcommunity2021gpt2spanish} GPT-2 checkpoints, gpt2-hindi \citep{surajp2021gpt2hindi},
Wenzhong-GPT2-110M \citep{zhang2022fengshenbang}, Qwen2.5-0.5B base and Instruct
\citep{qwen2025}, and SmolLM2-360M base and Instruct \citep{allal2025smollm2}.
Instruction-tuned checkpoints receive the same raw literary prefixes, without chat templates.
OLMo-2-1B \citep{olmo2}, whose released checkpoint OLMo-2-0425-1B has 1.48B parameters, extends the temperature grid in scale, and
Qwen2.5-3B-Instruct supplies next-token distributions on fixed contexts. English training text
is five books from NLTK's Gutenberg selection (Appendix~\ref{app:figvals}), with Austen's
\emph{Persuasion} held out, and
Appendix~\ref{app:multiwork} evaluates three further held-out works. German, Spanish and
Chinese use Project Gutenberg literature and Hindi uses Wikisource. The genres are not
uniform: the Spanish training text includes travel essays and the Chinese held-out text is a
travelogue (Appendix~\ref{app:figvals}).

\subsection{Ordered zero crossings across training temperatures}
\label{sec:crossings}

\paragraph{Design.} We sweep the training sampler's temperature and top-$p$ for English GPT-2
under full fine-tuning. A coarse grid of 15 cells (Table~\ref{tab:crossings}) is refined near
$T=1$ with five further temperatures at top-$p=1$ ($0.85$, $0.95$, $0.98$, $1.02$ and $1.05$).
Every cell has five seeds and nine generations, 900 trained generation-steps in all; corpus
sizes and optimizer settings are in Appendix~\ref{app:figvals}. The count analyses use a separate
eight-batch dump from the measurement sampler, while the logged $\Htwo$ series uses two batches.
Continuations are concatenated without a separator before counting; inserting separators changes
no conclusion (Appendix~\ref{app:repr}). A grid with generation batch 32 serves as a protocol
sensitivity check (Appendix~\ref{app:dense}).

For each order $\alpha$ we define $T^\ast(H_\alpha)$ as the first temperature at which the
seed-mean change from generation~0 to generation~9 crosses zero from below, using linear
interpolation between sampled cells. Crossings need not be unique: at top-$p=0.80$, richness
turns upward near $T=1.240$ and back down near $1.340$.

\paragraph{Refined crossings.} On the top-$p=1$ axis every order falls at $T=0.70$ and rises at
$T=1.25$, but each turns at a different temperature: $T^\ast(\Hinf)=0.927$,
$T^\ast(\Htwo)=0.962$, $T^\ast(H_1)=0.997$ and $T^\ast(\Hz)=1.035$. At four measured
temperatures ($0.95$, $0.98$, $1.00$ and $1.02$) richness falls while min-entropy rises; the
interval between the outer crossings is interpolated from these cells. We call
$T^\ast(\Hinf)<T^\ast(\Hz)$ the \emph{outer ordering} and the complete sequence
$T^\ast(\Hinf)<T^\ast(\Htwo)<T^\ast(H_1)<T^\ast(\Hz)$ the \emph{full ordering}. Resampling the
five seeds jointly across all cells (4,000 bootstrap draws) gives 95\% intervals of
$[0.917,0.950]$, $[0.959,0.963]$, $[0.993,1.001]$ and $[1.030,1.039]$, and every draw keeps the
full ordering. These intervals hold the grid fixed and so leave out interpolation error, which
is not negligible: refining the grid moved the $\Htwo$ root from $0.949$ to $0.961$ and then to
$0.962$, and the first move fell outside the coarse-grid interval of $[0.941,0.956]$. The third
decimal therefore describes the interpolant rather than the resolution of the experiment. On the coarse
grid, individual seeds follow the same ordering wherever all four roots are bracketed: in 5 of
5, 4 of 4 and 5 of 5 seeds at top-$p=1.0$, $0.8$ and $0.6$. The two top-$p=0.4$ cells bracket
no crossing. Across 18 sampled orders, all 4,000 common-length resamples preserve decreasing
crossing temperature with increasing order. Without interpolation, every measured temperature
has at most one change from negative to positive along those orders, in each seed and in the
mean (Appendix~\ref{app:posthocanalyses}). This single-crossing statement concerns only the
sampled orders; where signs differ, neither generation uniformly dominates the other on the
R\'enyi/Hill profile \citep{patil1982diversity,tothmeresz1995}. These analyses were not prespecified.

\paragraph{A fixed budget.} With common-length budgets, training at $T=1$ gives
$\Delta\Hz=-0.81$ bits (43\% fewer types) and $\Delta\Htwo=+0.94$ bits. Every grid corpus has
more than 13,000 words, so the analysis can be repeated at that fixed budget in every cell. At
$T=1$ it gives $\Delta\Hz=-0.400$ bits (24\% fewer types) and $\Delta\Htwo=+0.907$ bits, and
the roots become $0.936$, $0.963$, $0.990$ and $1.022$. The same four cells show falling
richness and rising min-entropy, and all 4,000 resamples keep the ordering. The budget changes
the size of the richness loss and the width of the interval. At tested prefixes of at most
1,000 words, however, generation~9 has more types at $T=1$ in every seed; the split appears
in four of five seeds at 5,000 words and all five at tested budgets from 8,000 upward.

\begin{table}[t]
\centering\small
\caption{Coarse 15-cell grid for English GPT-2 under full fine-tuning: five-seed means of the
change from generation~0 to generation~9 at common-length budgets ($48{,}125$--$70{,}367$
words). ``Types'' is the seed mean of the relative change in the number of word types, and
cross-perplexity is the generation-9 value relative to generation~0. The largest seed standard
deviation in the four entropy columns is $0.733$ bits ($\Hz$ at top-$p=0.40$, $T=1$); per-seed
values are released. Bold marks cells where richness and collision entropy move in opposite
directions, or where $\Delta\Htwo$ moves toward zero while cross-perplexity worsens.
Figure~\ref{fig:crossings} adds the five refinement temperatures.}
\label{tab:crossings}
\begin{tabular}{cc rrrr rr}
\toprule
top-$p$ & $T$ & $\Delta\Hz$ & $\Delta H_1$ & $\Delta\Htwo$ & $\Delta\Hinf$ & types & cross-ppl \\
\midrule
$1.00$ & $0.70$ & $-4.35$ & $-3.39$ & $-1.50$ & $-0.59$ & $-95\%$ & $11.6\times$ \\
$1.00$ & $0.90$ & $-2.91$ & $-2.26$ & $-0.89$ & $-0.23$ & $-87\%$ & $7.8\times$ \\
$1.00$ & $1.00$ & $\mathbf{-0.81}$ & $+0.08$ & $\mathbf{+0.94}$ & $+0.93$ & $\mathbf{-43\%}$ & $3.8\times$ \\
$1.00$ & $1.10$ & $+0.88$ & $+3.14$ & $+4.87$ & $+4.16$ & $+85\%$ & $8.1\times$ \\
$1.00$ & $1.25$ & $+1.44$ & $+4.36$ & $+6.62$ & $+6.66$ & $+172\%$ & $45.0\times$ \\
$1.00$ & $1.50$ & $+1.80$ & $+4.87$ & $+7.25$ & $+7.37$ & $+248\%$ & $109.8\times$ \\
\midrule
$0.80$ & $1.00$ & $-4.24$ & $-3.06$ & $-1.22$ & $-0.43$ & $-95\%$ & $22.4\times$ \\
$0.80$ & $1.09$ & $-3.57$ & $-2.52$ & $-0.76$ & $+0.10$ & $-92\%$ & $19.2\times$ \\
$0.80$ & $1.25$ & $+0.23$ & $+1.12$ & $+2.35$ & $+2.63$ & $+18\%$ & $14.1\times$ \\
$0.80$ & $1.50$ & $\mathbf{-0.41}$ & $+2.07$ & $\mathbf{+4.48}$ & $+5.30$ & $\mathbf{-23\%}$ & $98.9\times$ \\
\midrule
$0.60$ & $1.00$ & $-4.39$ & $-3.57$ & $-1.49$ & $-0.44$ & $-95\%$ & $11.6\times$ \\
$0.60$ & $1.18$ & $-4.45$ & $-3.27$ & $\mathbf{-1.26}$ & $-0.34$ & $-95\%$ & $\mathbf{37.3\times}$ \\
$0.60$ & $1.50$ & $+0.77$ & $+1.61$ & $+2.57$ & $+3.02$ & $+72\%$ & $34.7\times$ \\
\midrule
$0.40$ & $1.00$ & $-3.69$ & $-3.32$ & $-1.54$ & $-0.48$ & $-92\%$ & $8.1\times$ \\
$0.40$ & $1.30$ & $-4.63$ & $-3.49$ & $\mathbf{-1.41}$ & $-0.38$ & $-96\%$ & $\mathbf{24.9\times}$ \\
\bottomrule
\end{tabular}
\end{table}

\paragraph{Word frequencies behind the split.} At common length and $T=1$, \emph{the} falls from
4.9\% to 1.8\% of words. The ten commonest generation-0 types, all function words, account
for 104\% of the collision-probability decrease; the rest offset it slightly. Removing the
50 commonest types makes all four mean entropy changes negative
(Appendix~\ref{app:posthocanalyses}). This rise describes reduced function-word concentration
and could accompany grammatical degradation, which unigram counts do not test; reference
likelihood worsens here.

\paragraph{Apparent recovery while quality worsens.} At top-$p=0.60$, raising the training
temperature from $1.00$ to $1.18$ moves $\Delta\Htwo$ toward zero, from $-1.49$ to $-1.26$
bits, while cross-perplexity worsens from $11.6$ to $37.3$ times its generation-0 value; both
changes occur in all five seeds. Richness stays near $-4.4$ bits. A monitor watching $\Htwo$
alone would read this as partial recovery. At top-$p=0.80$, raising the temperature from $1.00$ to $1.09$ also moves $\Delta\Htwo$ toward zero, but there
cross-perplexity improves in four of five seeds.

\paragraph{Scope of the theory.} Tempering changes common and rare types at
different rates (Propositions~\ref{prop:temper} and~\ref{prop:sens}), and i.i.d.\ resampling
biases collision probability and observed support in known ways (Proposition~\ref{prop:rarefy}).
Higher-order plug-in entropy bias can have either sign. Predicting the crossing order through
nine neural refits under a fixed reader requires more than these one-step population and
sampling results. A
categorical simulation alternating tempering, 13,000 draws and refitting with uniform
smoothing of weight $0.02$, starting from our
measured unigram law, reproduces the outer ordering; the $\Hinf$ and $\Htwo$ roots remain in
the same grid bracket, and the neural roots remain empirical (Appendix~\ref{app:crossings}).

\paragraph{Scale, tuning and language.} Table~\ref{tab:xsize} repeats the top-$p=1$ temperature
axis for GPT-2 under LoRA, OLMo-2-1B under LoRA, German and Hindi GPT-2 under full fine-tuning,
and a Chinese model scored on characters. The outer ordering holds in every setting whose roots
are bracketed, and so does the full ordering at the interpolated roots. In Hindi, however, the
$\Hinf$ root is fragile. It sits where the seed-mean change at $T=1$ is $-0.0004$ bits against a
seed standard deviation of $0.64$, and stripping the Devanagari dandas that the primary tokenizer
leaves attached to words moves it from $1.000$ to $1.023$, just past the danda-stripped $\Htwo$ root ($1.021$; $1.020$ with the primary tokenizer),
a swap far smaller than the grid spacing. No measured Hindi cell shows the split; both outer
roots lie in $[1.0,1.1]$. Extending OLMo and Hindi from three to five seeds moves no
root by more than $0.02$. The interval sits in different places for different models: $T=1$ lies
inside it for fully fine-tuned GPT-2, OLMo and German, above it for GPT-2 LoRA, and at the lower
edge of its interpolated interval for Hindi.

Across eight representations of the English and Hindi texts, three transformations reproduce an
existing curve exactly. Of the 13 distinct measurements that remain, 12 keep the outer ordering
and 9 the full ordering (Appendix~\ref{app:repr}).

\begin{table}[t]
\centering\small
\caption{First upward crossings at top-$p=1$. Rows 1, 3 and 5 use five seeds and the others
three. Row~1 includes both refinements (crossing brackets of 0.02--0.05 in temperature); the
other grids are spaced 0.10--0.15 apart, so their third decimals are not resolved. The last column gives the interval between the outer crossings, rounded from the unrounded roots. In Hindi the $\Hinf$ root is fragile (see
text). The Han-only Chinese unit loses coverage before
most roots can be bracketed.}
\label{tab:xsize}
\begin{tabular}{llccccc}
\toprule
Setting & script / unit & $T^\ast(\Hz)$ & $T^\ast(H_1)$ & $T^\ast(\Htwo)$ & $T^\ast(\Hinf)$ & interval \\
\midrule
GPT-2 124M, full-FT & Latin / word & $1.035$ & $0.997$ & $0.962$ & $0.927$ & $[0.93,1.03]$ \\
GPT-2 124M, LoRA & Latin / word & $0.961$ & $0.950$ & $0.932$ & $0.884$ & $[0.88,0.96]$ \\
OLMo-2-1B ($1.48$B), LoRA & Latin / word & $1.020$ & $1.011$ & $1.000$ & $0.940$ & $[0.94,1.02]$ \\
\midrule
german-gpt2, full-FT & Latin / word & $1.055$ & $1.030$ & $1.004$ & $0.984$ & $[0.98,1.05]$ \\
gpt2-hindi, full-FT & Devanagari / word & $1.044$ & $1.035$ & $1.020$ & $1.000$ & $[1.00,1.04]$ \\
Wenzhong, full-FT & Han / \emph{char} & --- & --- & --- & $1.007$ & unit fails \\
\bottomrule
\end{tabular}
\end{table}

\paragraph{Script coverage in Chinese.} On the Chinese axis the share of Han characters among
non-whitespace characters falls from 85\% at generation~0 to 2.2\% at $T=1.10$ and 0.17\% at
$T=1.25$, where only 397 Han characters and 142 types remain, so we exclude that Han-only cell
for insufficient coverage. The other units keep their coverage but disagree with each other.
Scored on all non-whitespace characters at a common-length budget within each seed (about 20,000
characters), $\Delta\Htwo$ is negative at all five temperatures, from $-2.72$ to $-0.76$ bits;
at $T\ge1.10$ the output has shifted from Han to a small Latin alphabet. Scored on Wenzhong's own
tokens (about 44,000 per text), the same texts cross between $T=1$ and $1.10$, from $-0.17$ to
$+3.61$ bits. Which change is visible thus depends on both the unit's coverage and its
sensitivity to concentration, so coverage belongs next to the score; Appendix~\ref{sec:vocabdrift} relates
this script loss to the tokenizer's vocabulary (Table~\ref{tab:vocabdrift}). In English, the
four-order crossing sequence and the $T=1$ split survive matched top-$p=0.9$ and $1$ readers,
with top-$k$ disabled, at both word budgets (Appendix~\ref{sec:probes}).

\subsection{Distinct-\texorpdfstring{$n$}{n}, repetition and self-BLEU also cross}
\label{sec:notrenyi}

We rescore the 15 coarse cells with distinct-$n$, rep-4 and self-BLEU-4, comparing generation~9
with generation~0 at each pair's common-length budget over five seeds
(Table~\ref{tab:ngramcross}). Distinct-$n$ is the number of unique $n$-grams divided by the
number of $n$-gram positions, and rep-4 is one minus distinct-4. Self-BLEU-4 samples up to 40
newline-delimited segments of at least eight tokens as hypotheses and scores each against 10
other segments from the same capped text, using the closest reference length and add-one
smoothing for orders 2--4.

All four count metrics cross zero on the three well-sampled axes, top-$p=1.0$, $0.8$ and $0.6$.
Self-BLEU crosses at $1.0$ and $0.8$; at $0.6$ some seeds have too few valid segments for a
complete bracket. On the top-$p=1$ axis all five comparison metrics cross between $\Htwo$ and $\Hz$.
Phrase-frequency spectra differ: bigrams retain the full ordering but all roots lie below
$T=1$; trigrams reverse it, with $\Delta H_\infty=-0.83$ bits at $T=1$ as repeated phrases
concentrate the head (Appendix~\ref{app:posthocanalyses}).

At $T=1$ and top-$p=1$, the changes in distinct-2, distinct-3, rep-4 and self-BLEU are all
within $0.024$ of zero, and none has the same sign in all five seeds under this newline
segmentation. With fixed 50-token windows, self-BLEU falls in all five seeds. On the same texts $\Hz$
falls by $0.81$ bits and $\Htwo$ rises by $0.94$ bits, each with the same sign in every seed,
while cross-perplexity is $3.8$ times worse.

\begin{table}[t]
\centering\small
\caption{Coarse-grid comparison of generation~9 with generation~0 at each pair's common-length
budget, five-seed means, ordered by the top-$p=1$ crossing. Entropies are in bits; the other
scores lie in $[0,1]$, so their changes are not on a common scale. Bold marks untruncated
training at $T=1$, where four comparison metrics barely move and do not agree in sign across
seeds despite $3.8$ times worse cross-perplexity. Rep-4 and self-BLEU cross downward because
they rise with repetition. A dash marks too few valid segments for a complete five-seed
self-BLEU bracket; cell-level coverage is in the release.}
\label{tab:ngramcross}
\begin{tabular}{lrrrr@{\hspace{1.4em}}rrr}
\toprule
& \multicolumn{4}{c}{$\Delta$ at top-$p{=}1.0$} & \multicolumn{3}{c}{crossing $T^\ast$ at top-$p$} \\
\cmidrule(r){2-5}\cmidrule(l){6-8}
metric & $T{=}0.90$ & $T{=}1.00$ & $T{=}1.10$ & $T{=}1.50$ & $1.00$ & $0.80$ & $0.60$ \\
\midrule
$\Hinf$ & $-0.233$ & $+0.930$ & $+4.160$ & $+7.374$ & $0.920$ & $1.072$ & $1.213$ \\
$\Htwo$ & $-0.891$ & $+0.942$ & $+4.867$ & $+7.251$ & $0.949$ & $1.129$ & $1.285$ \\
self-BLEU-$4$ & $+0.197$ & $\mathbf{-0.019}$ & $-0.060$ & $-0.067$ & $0.991$ & $1.231$ & --- \\
distinct-$2$ & $-0.485$ & $\mathbf{+0.024}$ & $+0.335$ & $+0.337$ & $0.995$ & $1.207$ & $1.416$ \\
distinct-$3$ & $-0.572$ & $\mathbf{+0.008}$ & $+0.052$ & $+0.051$ & $0.999$ & $1.241$ & $1.485$ \\
rep-$4$ & $+0.460$ & $\mathbf{0.000}$ & $-0.006$ & $-0.006$ & $1.000$ & $1.249$ & $1.498$ \\
distinct-$1$ & $-0.131$ & $-0.065$ & $+0.131$ & $+0.392$ & $1.033$ & $1.222$ & $1.354$ \\
$\Hz$ & $-2.913$ & $-0.814$ & $+0.884$ & $+1.800$ & $1.048$ & $1.240$ & $1.453$ \\
\bottomrule
\end{tabular}
\end{table}

\subsection{Truncation lowers collision entropy relative to untruncated training}
\label{sec:robust}

An English study compares four training samplers: top-$p=1.0$, $0.95$ and $0.80$, and
top-$p=0.95$ with 10\% real text mixed into each training corpus. It starts from a GPT-2
checkpoint adapted to real text (Appendix~\ref{app:figvals}), whereas the temperature grid
starts from pretrained GPT-2. With top-$p=0.80$, word-level $\Htwo$ falls from $6.85$ to
$5.59$ bits while held-out cross-perplexity rises from $50.3$ to $2{,}189$, a $43.5$-fold
increase; the untruncated arm instead rises to $7.58$ bits (Figure~\ref{fig:collapse}). Mixing
in real text keeps the final cross-perplexity at $82.4$ ($1.6\times$), the least degraded of the
four arms, although the mixed arm also takes more optimizer steps because the added blocks train
under the same two-epoch schedule. On the truncated arms the models also grow more confident on
human text while their likelihood of that text falls (Appendix~\ref{sec:confquality}). A rerun also shows disagreement among standard diversity metrics (Table~\ref{tab:baselines}).

We repeat the truncated/untruncated comparison in nine further settings. They vary the language
(German, Spanish, Hindi and Chinese), the tuning regime (full fine-tuning or LoRA), the scale and
architecture, the checkpoint type (base or
instruction-tuned) and the unit (words or Han characters). Table~\ref{tab:robust} reports
generations 0 and 9 for both arms.

\begin{table}[t]
\centering\small
\caption{Ten self-training settings: three-seed means under the measurement sampler, with a
shared generation~0. The $\Htwo$ columns use the logged measurement samples at full length; type
changes use the first 13,000 units of separate dumps (Table~\ref{tab:dissociate}), because
richness depends strongly on length. Arrows give the sign of the mean change from generation~0. The English row is an independent
rerun of the four-arm study's truncated and untruncated arms; its separate dumps give $5.65$ and
$7.46$ bits in Table~\ref{tab:baselines}, reflecting different measurement samples.
On an English generation-0 text (the seed-0 baseline of the main grid), i.i.d.\ resampling at
$N=13{,}000$ gives a $0.013$-bit $\Htwo$ bias (Appendix~\ref{app:rarefy-emp}); this is not a
general correction. Character and word values have different scales. $^\ast$Sampled with the
checkpoint's default repetition penalty of 1.1; without it the truncated arm falls to $5.54$ bits
(\S\ref{sec:richness}).}
\label{tab:robust}
\begin{tabular}{llcccc}
\toprule
 & & & \multicolumn{2}{c}{generation-9 $\Htwo$} & \\
\cmidrule(lr){4-5}
Setting & unit & gen.~0 & truncated & untruncated & types (trunc./untrunc.) \\
\midrule
English full-FT, GPT-2 & word & $6.85$ & $5.62\downarrow$ & $7.48\uparrow$ & $-89\%$ / $-29\%$ \\
German full-FT, GPT-2 & word & $7.51$ & $5.67\downarrow$ & $7.65\uparrow$ & $-92\%$ / $-32\%$ \\
Spanish full-FT, GPT-2 & word & $6.32$ & $4.80\downarrow$ & $6.39\uparrow$ & $-90\%$ / $-10\%$ \\
English LoRA, GPT-2 & word & $6.97$ & $5.02\downarrow$ & $8.07\uparrow$ & $-92\%$ / $+46\%$ \\
English LoRA, Qwen-0.5B & word & $6.61$ & $5.50\downarrow$ & $6.71\uparrow$ & $-83\%$ / $-22\%$ \\
English LoRA, Qwen-Instruct$^\ast$ & word & $7.62$ & $8.13\uparrow$ & $9.82\uparrow$ & $-59\%$ / $+36\%$ \\
English LoRA, SmolLM2-Instruct & word & $6.69$ & $4.88\downarrow$ & $6.73\uparrow$ & $-88\%$ / $-15\%$ \\
English LoRA, SmolLM2-base & word & $6.66$ & $4.67\downarrow$ & $6.69\uparrow$ & $-88\%$ / $-17\%$ \\
\textbf{Hindi} full-FT, gpt2-hindi & word & $7.26$ & $5.07\downarrow$ & $7.10\downarrow$ & $-91\%$ / $-35\%$ \\
\textbf{Chinese} full-FT, Wenzhong & char & $8.20$ & $6.28\downarrow$ & $7.02\downarrow$ & $-78\%$ / $-58\%$ \\
\bottomrule
\end{tabular}
\end{table}

\paragraph{Script and segmentation.} Hindi gives a non-Latin, word-tokenized comparison. Its
truncated arm falls from $7.26$ to $5.07$ bits and loses 91\% of its observed types, while the
untruncated arm ends at $7.10$ bits. The comparison with Chinese also changes model depth
(six layers in Hindi, twelve in Chinese), corpus, tokenizer and measurement unit. Hindi's untruncated arm loses a little collision entropy while gaining min-entropy, so Hindi meets the sign criterion of
\S\ref{sec:sweep} only at the two highest orders. The Hindi data
are Premchand works from Wikisource (Appendix~\ref{app:figvals}).

In all ten settings and at all nine R\'enyi orders of \S\ref{sec:sweep}, the three truncated
seeds lie below all three untruncated seeds. The mean paired gap ranges from $0.40$ to $4.93$
bits (median $1.75$), no arm's seed standard deviation of the change exceeds $0.70$
bits, and all 270 paired
seed differences have the same sign, the smallest being $0.27$ bits. With three seeds these
statements are descriptive. Cross-perplexity rises $2.5$- to $48$-fold on the truncated arms,
against at most $4.7$-fold on the untruncated ones, and low-rank tuning does not by itself limit
the damage: across the LoRA settings the truncated arms' cross-perplexity rises $2.5$- to
$21.8$-fold.

\subsection{Separation by the direction of change}
\label{sec:sweep}

Magnitude separates the arms, but a monitor often sees one training trajectory at a time and
would like the direction of change to say which kind of drift is under way. We therefore ask
whether the arms move in \emph{opposite} directions: every truncated seed falls and every
untruncated seed rises, or the reverse. Table~\ref{tab:sweep} counts the settings that meet this
criterion at nine orders, $\alpha\in\{0,\tfrac12,1,\tfrac32,2,3,5,10,\infty\}$.

\begin{table}[t]
\centering\small
\caption{Settings, out of ten, in which all truncated seeds move one way and all untruncated
seeds the other, by R\'enyi order. Changes below $0.02$ bits decide some entries, so individual
classifications may not be stable across reruns. This sign criterion is stricter than the
magnitude separation of \S\ref{sec:robust}. The counts include the penalized
Qwen2.5-0.5B-Instruct setting (\S\ref{sec:richness}), which meets the criterion at $\alpha\le1.5$;
with its penalty-free rerun instead, they are 1, 1, 2, 2, 3, 7, 7, 8 and 7.}
\label{tab:sweep}
\begin{tabular}{lccccccccc}
\toprule
$\alpha$ & $0$ & $0.5$ & $1$ & $1.5$ & $2$ & $3$ & $5$ & $10$ & $\infty$ \\
\midrule
settings meeting the criterion & 2 & 2 & 3 & 3 & 3 & 6 & 6 & \textbf{7} & \textbf{7} \\
\bottomrule
\end{tabular}
\end{table}

Direction separates the arms in at most seven of ten settings, at $\alpha=10$ and $\infty$
(eight at $\alpha=10$ with the penalty-free Qwen-Instruct rerun); in
six at $\alpha=3$ and $5$; and in three at $\alpha=2$. Higher orders emphasize common types;
locating degradation within the frequency distribution requires evidence beyond arm separation. Many failures come
from untruncated arms that barely move: at the fixed 13,000-word budget their mean $\Htwo$
changes are $+0.02$ (German), $-0.03$ (Qwen base; $+0.10$ in Table~\ref{tab:robust}'s logged samples), $+0.08$ (Spanish), $+0.09$
(SmolLM2-Instruct) and $+0.05$ (SmolLM2-base) bits. These arms show little over-dispersion; a
high-temperature arm tests that condition directly (Appendix~\ref{sec:threepoint}). A weaker
version of the split between richness and concentration does appear in five of the ten
untruncated arms (English, German, Spanish and both SmolLM2 checkpoints): at the fixed budget they
lose 10--32\% of their types while their $\Htwo$ holds or rises ($\Delta\Htwo\ge0$). Appendix~\ref{app:rarefy-emp}
gives sample-size and seed-variance controls, and Table~\ref{tab:dissociate} the per-setting
profiles.

\subsection{An exception traced to a sampler setting}
\label{sec:richness}

In Qwen2.5-0.5B-Instruct under LoRA, the truncated arm loses 59\% of its types ($\Hz$ falls by
$1.30$ bits) while $\Htwo$ rises by $0.47$ bits and $\Hinf$ by $1.62$, and cross-perplexity is
22 times worse (Tables~\ref{tab:dissociate} and~\ref{tab:qwenrp}). The text has fewer types but a less concentrated
frequency distribution, so rising collision entropy coexists here with both lost richness and
worse reference likelihood. SmolLM2-360M-Instruct, from a different family, contracts normally at
every order. On the Qwen checkpoint, sampled with the repetition penalty discussed below, the
effect holds across truncation levels: at the fixed 13,000-word budget $\Htwo$ rises in every seed at every tested top-$p$ from $0.60$ to $1.00$,
though only by $0.07$ bits at $0.60$. Leakage of other scripts does not explain it:
non-ASCII words, mostly Chinese and Cyrillic, make up $0.6\%$ of the first 13,000 words at
generation~0 and $0.5\%$ on the truncated arm at generation~9.

Removing the inherited repetition penalty eliminates this exception. The checkpoint sets it to
1.1, inherited in every phase (\S\ref{sec:setup}), down-weighting tokens already in the context.
We reran both arms and the checkpoint's three other training samplers at penalty 1.0, matching
initial adapters, random streams, corpus and budgets (Table~\ref{tab:qwenrp}). Without the penalty, generation-0 collision entropy is $6.57$
bits, close to the base model's, and the truncated arm contracts like every other setting: at
13,000 words it loses 86\% of its types, $\Hz$, $H_1$ and $\Htwo$ fall by $2.84$, $2.67$ and
$1.18$ bits ($\Htwo$ in all three seeds), and $\Hinf$ falls by $0.19$ on average. The untruncated
arm shows the weaker split of the other untruncated arms (\S\ref{sec:sweep}), losing 26\% of its types while $\Htwo$ holds ($+0.02$) and $\Hinf$ rises ($+0.43$). The
top-$p=0.60$ and $0.95$ arms now contract as well, with $\Htwo$ falling by $1.53$ and $0.79$ bits
in every seed, while training at $T=1.5$ still raises it. The other tables and detection pool retain the labeled penalized runs: they show that
another sampler setting can produce the same richness--concentration disagreement.

\subsection{What the profile can and cannot diagnose}
\label{sec:detect}

We test profile movement, discrimination and threshold transfer. A run is labeled degraded
when held-out cross-perplexity reaches twice its generation-0 value; this measures fit to one
reference text. Appendix~\ref{app:detect} gives protocols, controls and an intervention study.

\paragraph{Movement and sign.} Let $M_g=\max(|\Delta\Hz|,|\Delta\Hinf|)$ at generation $g$,
and let $D_g$ indicate that the two changes have opposite signs and both exceed $0.20$ bits in
magnitude, a floor chosen to reduce sensitivity to sampling noise. The development pool has 95
trajectories above the cutoff, the 75 coarse-grid runs and 20 Pythia-160m runs from the first
stage of the frozen-detector study (Appendix~\ref{app:detect}), together with six no-training
controls and three runs adapted on human text. The trajectory score $\max_g M_g$ reaches AUROC $0.994$, but six of its nine
negatives never trained, so this mainly shows that it detects movement. The three alarm rules behave differently (Table~\ref{tab:alarms}, Appendix~\ref{app:detect}):
magnitude alone fires on 92 of 95 above-cutoff runs, and adding the sign clause catches all 95.
Without its floor the clause also fires on every unchanged model, and even with the floor it flags all
three human-data adaptations, whose profiles change by $\Delta\Hz=-0.636$ and
$\Delta\Hinf=+0.433$ bits at 13,000 words while cross-perplexity improves to $0.60$ times its
starting value. These controls use top-$k=50$; the alarm also flags all five human adaptations
under the primary protocol (Appendix~\ref{app:freshcontrols}).

\paragraph{Cross-setting classification.} On 660 corpora from 16 settings, we choose a feature
pair by nested leave-one-setting-out validation and score it on the held-out setting. The mean
AUROC over the 14 folds that contain both classes is $0.897$. Selection picks
$(\Htwo,\text{distinct-}2)$ in 11 folds, $(\Hz,\Htwo)$ in two and $(\Hz,H_1)$ in one. Evaluated
as a fixed pair in every fold, $(\Htwo,\text{distinct-}2)$ has the highest mean of the 15 pairs,
$0.963$, against $0.961$ for $(\Hz,\Htwo)$ and $0.917$ for $(\Hz,H_1)$; differences this small
do not single out a best pair. A one-feature post-hoc baseline exceeds both leading fixed pairs: unsigned richness change $|\Delta\Hz|$
scores $0.970$, while generation index alone, which uses no text, scores $0.908$.
The nested and fixed-pair numbers evaluate different procedures,
so their gap is not an estimate of selection bias. The worst fold is English $T=2$, where nested
selection chooses $(\Hz,H_1)$ and scores $0.194$. Appendices~\ref{app:posthocanalyses} and~\ref{app:familyholdout} give the in-sample fit and family holdouts. Four settings are
Qwen2.5-0.5B-Instruct runs sampled with the checkpoint's repetition penalty.

\paragraph{Pythia: one training regime, three decoders.} Pythia-160m \citep{biderman2023pythia}
is trained with five seeds at each of three decoders under a single regime: pure recursion,
learning rate $10^{-5}$ and one epoch. The $2\times$ cutoff on \emph{Persuasion} splits the
untruncated $T=1$ arm, with three seeds below it ($1.84$, $1.93$ and $1.94$) and two above
($2.07$ and $2.09$), and all ten runs at top-$p=0.8$ or $T=2$ (top-$p=0.95$) exceed it. The
exploratory endpoint score
$\max(|\Delta\Hz|,|\Delta\Htwo|)$ has AUROC $0.944$, the same as richness alone. That figure
rests on 30 of 30 concordant comparisons across decoders but only 4 of 6 within the $T=1$ arm
(AUROC $0.667$), so the pooled ranking mostly separates decoders rather than quality within a
decoder. One-sided scores stay near chance ($0.472$--$0.528$). Appendix~\ref{app:detect}
describes a preliminary frozen-detector test whose healthy runs used a different training recipe.

\paragraph{Changing the reference work.} With models, scores and cutoff held fixed, three more
books confirm that the GPT-2 grid loses likelihood: all 100 endpoints exceed the $2\times$
cutoff on every work, and at $T=1$ the mean ratios are $3.43$--$4.48$, against $3.83$ on
\emph{Persuasion}. Pythia's labels move more. \emph{Sense and Sensibility} places 11 of 15 runs
above the cutoff, while \emph{Moby Dick}, \emph{The Adventures of Buster Bear} and the three-book
aggregate place all 15 above it, which leaves no negative class and makes AUROC undefined
(Table~\ref{tab:multiwork}). Even within English literature, the choice of reference text changes the labels.

\section{Conclusion}
\label{sec:conclusion}

On the English GPT-2 training-temperature grid, min-entropy changes sign first and richness
last. Between these crossings, models have fewer word types and a less dominant most-common
word. At sufficient budgets, $T=1$ lies inside this interval under all three readers. The full
ordering holds at interpolated roots in every bracketed setting (Hindi has no measured split
cell), but in only 9 of 13 distinct measurements under eight representations; 12 keep the outer
ordering. The interval moves with the setting, and phrase-level spectra can reverse the order.

Report richness and higher orders at a stated budget and representation, with independent
quality measures. The same split occurs during human-text fine-tuning while likelihood improves,
so the signs of entropy changes do not identify degradation or its cause. Validate thresholds on healthy trajectories
under the intended decoder and reference text.

\section*{Limitations}

\paragraph{Theory and resolution.} The theory characterizes decoding and sampling effects;
the neural crossing order remains empirical. Most comparisons use three to five seeds;
coarse grids and linear interpolation limit precision, and bootstrap intervals exclude interpolation error. The Hindi
min-entropy root is fragile, and the vocabulary-composition comparison of
Appendix~\ref{sec:vocabdrift} is a heuristic based on three settings.

\paragraph{Quality and transfer.} Quality is measured only through reference likelihood and
automatic metrics, without human judgments. The other languages each use a single held-out work,
and we do not know which works the models saw in pretraining. Setting-level folds share model
families, Pythia's negative class depends on the reference, and the intervention results rest on
five seeds, so neither threshold transfer nor intervention timing is validated.

\paragraph{Protocol and scale.} The recursive models have at most $1.48$B parameters and train on
literary corpora of uneven genre. Because most settings train for two epochs per generation, the
number of updates changes with corpus length and with added real data. EOS handling differs by tokenizer,
and the GPT-2 EOS variant also halves the generation batch (Appendix~\ref{app:repr}). The
Qwen-Instruct runs in the main tables inherited a repetition penalty, which only the reruns of
\S\ref{sec:richness} remove. Training on Apple MPS is not bitwise deterministic.

\paragraph{Measurement and data.} Word, character and model-token entropies are not comparable,
a Han-only unit can lose almost all coverage, and unigram measures miss phrase-level loops. The
released count multisets recompute R\'enyi and $n$-gram metrics but not the 12,000-word detection
features (released as rows), self-BLEU, MAUVE or perplexity. Released endpoint, reader and control texts support reanalysis. The counts cover every
setting except two whose text was not kept (the four-arm study and the temperature sweep of
Appendix~\ref{sec:tempfails}). Several secondary runs lack exact revision or resume provenance (Appendix~\ref{app:figvals});
the endpoint MAUVE featurizer is also not fully recorded (Appendix~\ref{app:detect}).

\section*{Broader Impact Statement}
Treating a fall in diversity as evidence of poor quality can penalize legitimate specialization
or underrepresented varieties, so these diagnostics should be used with explicit units, coverage
checks and an independent quality evaluation before they filter data or models.

\section*{Use of AI Assistants}
Claude Code and OpenAI Codex assisted with drafting and revising the text, writing and running
experiment code, and checking derivations, references and numerical results; the release
documents these checks. The author directed the research and is responsible for the paper's
claims, interpretation and submission.

\bibliography{custom}
\bibliographystyle{tmlr}

\clearpage
\appendix

\section{Supporting experiments}
\label{app:support}

This appendix collects the per-setting profiles behind \S\ref{sec:robust} and
\S\ref{sec:richness} (Table~\ref{tab:dissociate}), the repetition-penalty control
(Table~\ref{tab:qwenrp}), standard diversity metrics on a rerun of the four-arm study's truncated
and untruncated arms (Table~\ref{tab:baselines}), and analyses of predictive confidence, training
temperature, the measurement sampler, high-temperature training and the tokenizer's script
composition.

\begin{table}[ht]
\centering\small
\caption{Change in $H_\alpha$ from generation~0 to generation~9 on the truncated arm: three-seed
means at the fixed 13,000-unit budget, which every setting meets. Richness ($\Hz$) falls everywhere.
On the instruction-tuned Qwen checkpoint, sampled with its default repetition penalty of 1.1,
the higher orders move the other way: the model restricts its vocabulary while using the
remaining words more evenly. $^\ast$Without the penalty all four orders fall
(Table~\ref{tab:qwenrp}).}
\label{tab:dissociate}
\begin{tabular}{lrrrr}
\toprule
Setting (truncated arm) & $\Delta\Hz$ & $\Delta H_1$ & $\Delta\Htwo$ & $\Delta\Hinf$ \\
\midrule
English full-FT, GPT-2 & $-3.25$ & $-2.49$ & $-1.23$ & $-0.45$ \\
German full-FT, GPT-2 & $-3.59$ & $-3.46$ & $-1.97$ & $-1.14$ \\
Spanish full-FT, GPT-2 & $-3.30$ & $-3.44$ & $-1.46$ & $-0.27$ \\
English LoRA, GPT-2 & $-3.60$ & $-3.59$ & $-1.96$ & $-0.55$ \\
English LoRA, Qwen-0.5B & $-2.55$ & $-2.45$ & $-1.25$ & $-0.32$ \\
English LoRA, SmolLM2-Instruct & $-3.04$ & $-3.12$ & $-1.67$ & $-0.40$ \\
English LoRA, SmolLM2-base & $-3.06$ & $-3.22$ & $-1.84$ & $-0.55$ \\
\textbf{Hindi} full-FT, gpt2-hindi & $-3.41$ & $-4.15$ & $-2.44$ & $-1.19$ \\
\textbf{Chinese} full-FT, Wenzhong & $-2.18$ & $-2.29$ & $-1.71$ & $-0.42$ \\
\textbf{English LoRA, Qwen-Instruct}$^\ast$ & $\mathbf{-1.30}$ & $\mathbf{-0.95}$ & $\mathbf{+0.47}$ & $\mathbf{+1.62}$ \\
\bottomrule
\end{tabular}
\end{table}

\begin{table}[ht]
\centering\small
\caption{Qwen2.5-0.5B-Instruct under LoRA with the repetition penalty of 1.1 inherited from the
checkpoint's generation configuration, and with the penalty set to 1.0, for each training sampler
(top-$p$ at $T=1$, or $T=1.5$ at top-$p=0.95$). Everything else is matched: initial adapters,
random streams, corpus and budgets. Three-seed means. The two $\Htwo$ columns are the logged two-batch
measurement samples; the changes $\Delta$ and ``types'' (the change in the number of word types)
use the separate eight-batch dumps and compare generation~9 with generation~0 at 13,000 words;
cross-perplexity is the generation-9 ratio to generation~0. The penalized top-$p$ 0.80 and 1.0
rows are the truncated and untruncated arms of Table~\ref{tab:robust}. $^\dagger$Generation-9 dumps
are shorter than 13,000 words, so each seed's changes use the common length of its generation-0
and generation-9 dumps (10,369--11,096 words). About 94\% of their words are distinct, so the
logged value, from a sample about a quarter as long, sits near $\log_2$ of that sample's length
and differs from the dump-based endpoint.}
\label{tab:qwenrp}
\setlength{\tabcolsep}{4.5pt}
\begin{tabular}{llcccccccc}
\toprule
penalty & sampler & $\Htwo$, gen.~0 & $\Htwo$, gen.~9 & $\Delta\Hz$ & $\Delta H_1$ & $\Delta\Htwo$ & $\Delta\Hinf$ & types & cross-ppl \\
\midrule
1.1 & top-$p$ 0.60 & $7.62$ & $7.70$ & $-1.58$ & $-1.32$ & $+0.07$ & $+1.12$ & $-66\%$ & $16.4\times$ \\
1.1 & top-$p$ 0.80 & $7.62$ & $8.13$ & $-1.30$ & $-0.95$ & $+0.47$ & $+1.62$ & $-59\%$ & $21.8\times$ \\
1.1 & top-$p$ 0.95 & $7.62$ & $8.85$ & $-0.45$ & $+0.06$ & $+1.29$ & $+2.42$ & $-27\%$ & $5.5\times$ \\
1.1 & top-$p$ 1.0 & $7.62$ & $9.82$ & $+0.44$ & $+1.34$ & $+2.35$ & $+2.90$ & $+36\%$ & $3.4\times$ \\
1.1 & $T=1.5^\dagger$ & $7.62$ & $11.32$ & $+1.60$ & $+3.44$ & $+5.48$ & $+5.68$ & $+203\%$ & $3.0\times$ \\
\midrule
1.0 & top-$p$ 0.60 & $6.57$ & $5.08$ & $-3.20$ & $-2.96$ & $-1.53$ & $-0.40$ & $-89\%$ & $2.2\times$ \\
1.0 & top-$p$ 0.80 & $6.57$ & $5.54$ & $-2.84$ & $-2.67$ & $-1.18$ & $-0.19$ & $-86\%$ & $10.8\times$ \\
1.0 & top-$p$ 0.95 & $6.57$ & $5.75$ & $-1.80$ & $-1.99$ & $-0.79$ & $+0.17$ & $-71\%$ & $3.0\times$ \\
1.0 & top-$p$ 1.0 & $6.57$ & $6.68$ & $-0.44$ & $-0.56$ & $+0.02$ & $+0.43$ & $-26\%$ & $1.5\times$ \\
1.0 & $T=1.5^\dagger$ & $6.57$ & $11.31$ & $+1.87$ & $+4.11$ & $+6.52$ & $+6.57$ & $+265\%$ & $2.9\times$ \\
\bottomrule
\end{tabular}
\end{table}

\begin{table}[ht]
\centering\small
\caption{Standard diversity metrics on the rerun of the four-arm study's truncated and untruncated
arms (the English row of Table~\ref{tab:robust}), generation~0 to~9, three-seed means over
full-length texts (70,106--77,122 words at generations 0 and 9). A metric flags the untruncated arm when it changes by at least 10\% in the
conventional collapse direction, an exploratory rule. Self-BLEU here is sentence BLEU-2 with
NLTK's method-1 smoothing on the first 80 eligible sentences, a different protocol from the
grid's BLEU-4. At 13,000 words (Table~\ref{tab:robust}), the untruncated arm loses 29\%
of its types, while its rising $\Htwo$ reflects more even use of a smaller observed vocabulary.}
\label{tab:baselines}
\begin{tabular}{lccc}
\toprule
Metric & truncated (top-$p$ $0.80$) & untruncated (top-$p$ $1.0$) & flags untruncated arm? \\
\midrule
$\Htwo$ & $6.86\!\to\!5.65\,\downarrow$ & $6.86\!\to\!7.46\,\uparrow$ & no \\
Shannon $H_1$ & $9.14\!\to\!6.50\,\downarrow$ & $9.14\!\to\!9.12\,\approx$ & no \\
distinct-1 & $0.086\!\to\!0.006\,\downarrow$ & $0.086\!\to\!0.048\,\downarrow$ & \textbf{yes} (gen.~2) \\
distinct-2 & $0.55\!\to\!0.04\,\downarrow$ & $0.55\!\to\!0.53\,\approx$ & no \\
self-BLEU & $0.44\!\to\!0.93\,\uparrow$ & $0.44\!\to\!0.39\,\downarrow$ & no \\
MAUVE & $0.64\!\to\!0.10\,\downarrow$ & $0.64\!\to\!0.46\,\downarrow$ & \textbf{yes} (gen.~2) \\
\bottomrule
\end{tabular}
\end{table}

\begin{figure}[ht]
\centering
\includegraphics[width=\linewidth]{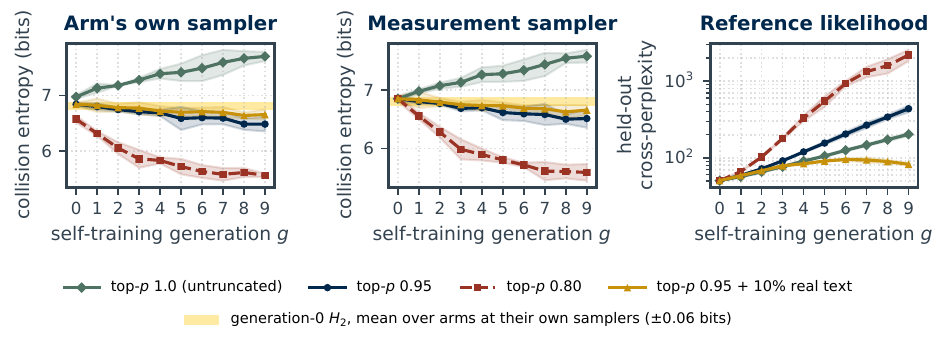}
\caption{The four-arm English study with common random numbers, three seeds. Truncated
training samplers lower word-level $\Htwo$ from generation to generation, while untruncated
training (top-$p=1$) drifts \emph{up}. Held-out cross-perplexity rises fastest for the tightest
nucleus, and mixing 10\% real text into each training corpus limits the rise to $1.6\times$.
The shaded band in the two $\Htwo$ panels marks the generation-0 value read with the arms' own
samplers, averaged over arms and seeds ($6.81$ bits); its $\pm0.06$-bit width is for display only.}
\label{fig:collapse}
\end{figure}

\subsection{Predictive confidence and reference likelihood}
\label{sec:confquality}

\begin{figure}[ht]
\centering
\includegraphics[width=\linewidth]{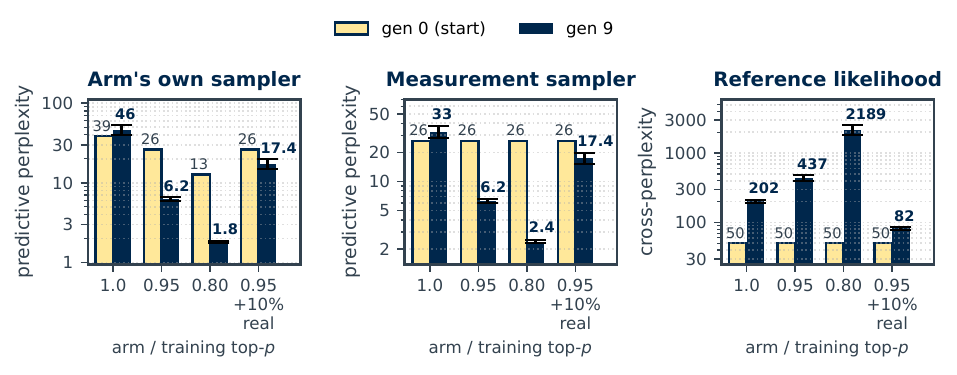}
\caption{English GPT-2, generation 0 to 9. On the truncated arms, predictive perplexity
$2^{\overline{H_1(q_t)}}$ falls while cross-perplexity on held-out real text rises. The first
two panels average entropy over human contexts, using the arm's own sampler or the measurement
sampler; they do not score generated sequences. The rightmost group in each panel is the
top-$p=0.95$ arm with 10\% real text.}
\label{fig:selfcross}
\end{figure}

We measure predictive perplexity as $2^{\overline{H_1(q_t)}}$, the exponentiated mean entropy of
the truncated next-token law $q_t$ over held-out human contexts. It uses no generated samples and
differs both from model-scored sample perplexity and from word-level unigram entropy.
Cross-perplexity instead scores the observed reference tokens under the untruncated model. On the
top-$p=0.80$ arm, predictive perplexity under the measurement sampler falls from $26.3$ to $2.4$
while cross-perplexity rises from $50.3$ to $2{,}189$: more concentrated predictions accompany
worse reference likelihood (Figure~\ref{fig:selfcross}).

\subsection{Decoder and training temperature sweeps}
\label{sec:tempfails}

\begin{figure}[ht]
\centering
\includegraphics[width=\linewidth]{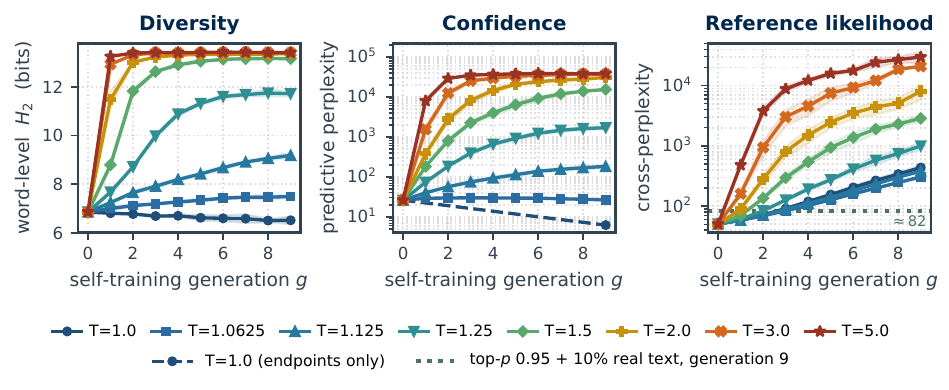}
\caption{High-temperature self-training (English GPT-2, training top-$p$ fixed at $0.95$,
$T=1$--$5$; $\Htwo$ read with the measurement sampler; three seeds). \emph{Left:} high
training temperatures raise $\Htwo$. \emph{Centre:} predictive perplexity
$2^{\overline{H_1(q_t)}}$ on human contexts rises at high training temperatures; the $T=1$
series has endpoint measurements only and is dashed. \emph{Right:} every endpoint is worse than
generation~0, with larger increases at high $T$; the dotted line is the generation-9
cross-perplexity of the top-$p=0.95$ arm with 10\% real text (about 82).}
\label{fig:tempbracket}
\end{figure}

Propositions~\ref{thm:temp} and~\ref{thm:trunc} give opposing next-token effects, and
Proposition~\ref{prop:cap} bounds entropy only while the retained set stays fixed, so it gives no
document-level bound when contexts have different nuclei. Holding the training top-$p$ at $0.95$
and sweeping the \emph{training} temperature on GPT-2, generation-9 $\Htwo$ covers the whole
range: $6.5$ bits at $T=1$, $7.5$ at $T=1.0625$, $9.2$ at $T=1.125$ and $13.2$--$13.4$
for $T\ge1.5$. Tightening truncation instead moves it the other way, to $5.6$. The two knobs thus
push $\Htwo$ in opposite directions from a shared start. Quality is worse than at generation~0 ($50.3$) everywhere on the sweep, though not monotonically:
cross-perplexity is $437$, $306$, $381$, $999$, $2{,}860$, $8{,}072$, $20{,}923$ and $29{,}844$
at $T=1$, $1.0625$, $1.125$, $1.25$, $1.5$, $2$, $3$ and $5$. In this comparison mixing in real data gives a smaller
perplexity increase and keeps diversity near its starting range (Figure~\ref{fig:tempbracket}).

\subsection{Changing the measurement sampler}
\label{sec:probes}

\paragraph{Matched readers.} To test whether the split depends on measurement truncation,
we reread all 55 retained generation-9 checkpoints on the eleven-temperature English top-$p=1$
training axis, with measurement top-$p=1$ and $0.9$, both at $T=1$ and top-$k=0$.
Each reader gets five new generation-0 baselines and 55 endpoints (120 corpora total).
We keep the original prompt pool, eight batches of 64 prefixes of 16 tokens, up to 176 new
tokens per continuation, boundary handling and decoding, resetting the seed
$100000s+1000g+4$ separately for each reader. Each endpoint is compared with its own reader's
baseline at common length (60,694--71,302 words across these pairs) and at 13,000 words.
This post-hoc check reuses the trained seeds; the top-$p=0.95$ rows repeat the primary measurement.

\begin{table}[ht]
\centering\small
\setlength{\tabcolsep}{4pt}
\caption{Reader sensitivity on the eleven-temperature English grid (five seeds).
All readers use $T=1$ and top-$k=0$. The $p=0.95$ rows repeat the
primary measurement; $p=0.9$ and $1$ are matched readings of the same checkpoints. Roots are
first upward interpolations of the seed mean. Changes at training $T=1$ are in bits;
``split'' counts seeds with $\Delta\Hz<0<\Delta\Hinf$.}
\label{tab:readers}
\begin{tabular}{llccccccc}
\toprule
budget & reader $p$ & $T^\ast(\Hinf)$ & $T^\ast(\Htwo)$ & $T^\ast(H_1)$ & $T^\ast(\Hz)$ & $\Delta\Hz$ & $\Delta\Hinf$ & split \\
\midrule
common & $0.9$ & $0.924$ & $0.959$ & $0.993$ & $1.029$ & $-0.746$ & $0.932$ & 5/5 \\
common & $0.95$ & $0.927$ & $0.962$ & $0.997$ & $1.035$ & $-0.814$ & $0.930$ & 5/5 \\
common & $1$ & $0.929$ & $0.964$ & $1.003$ & $1.041$ & $-0.882$ & $0.951$ & 5/5 \\
\midrule
$13{,}000$ & $0.9$ & $0.925$ & $0.959$ & $0.984$ & $1.016$ & $-0.317$ & $0.911$ & 5/5 \\
$13{,}000$ & $0.95$ & $0.936$ & $0.963$ & $0.990$ & $1.022$ & $-0.400$ & $0.929$ & 5/5 \\
$13{,}000$ & $1$ & $0.939$ & $0.964$ & $0.996$ & $1.029$ & $-0.499$ & $0.943$ & 5/5 \\
\bottomrule
\end{tabular}
\end{table}

The four mean roots retain $T^\ast(\Hinf)<T^\ast(\Htwo)<T^\ast(H_1)<T^\ast(\Hz)$
under measurement top-$p=0.9$ and $1.0$ at both budgets (Table~\ref{tab:readers}). In each of these four comparisons,
all 4,000 joint seed-bootstrap draws have all four roots and keep the strict ordering.
The released percentile intervals condition on the measured grid and exclude interpolation error.
At training $T=1$, every seed has $\Delta\Hz<0<\Delta\Hinf$ under both readers and budgets.
The profile still depends on the reader: common-length $\Delta H_1$ at $T=1$ is $-0.128$ bits
with top-$p=1$, versus $+0.221$ with top-$p=0.9$; at 13,000 words these become $+0.120$ and
$+0.423$. The outer split and crossing order persist across these readers, while the
Shannon sign depends on reader and budget.
All texts, checkpoint and source hashes, per-seed results and a CPU reconstruction script are released.

\paragraph{Additional measurement protocols.}
The main grid is read with top-$p=0.95$ and $T=1$, with top-$k$ disabled. A secondary analysis
reads its thirty top-$p=1$ endpoint checkpoints in three other ways (Table~\ref{tab:probes}): with
two sampled measurements, top-$p=0.95$ and top-$p=1$, both at $T=1$ and both with the library's default
top-$k=50$; and with teacher-forced next-token
distributions on human text, averaging $H_\alpha$ of the full softmax over positions without
sampling.

\begin{table}[ht]
\centering\small
\caption{Crossing temperatures on the top-$p=1$ axis under alternative measurements of the same
endpoint checkpoints. Teacher-forced $\Hz$ is constant at $\log_2|\V|$ for a dense softmax, so it
has no crossing.}
\label{tab:probes}
\begin{tabular}{lccccc}
\toprule
measurement & $T^\ast(\Hz)$ & $T^\ast(H_1)$ & $T^\ast(\Htwo)$ & $T^\ast(\Hinf)$ & ordered in $\alpha$ \\
\midrule
top-$p=0.95$, top-$k=50$, $T=1$ & $1.010$ & $0.955$ & $0.937$ & $0.928$ & yes \\
top-$p=1$, top-$k=50$, $T=1$ & $1.009$ & $0.955$ & $0.934$ & $0.913$ & yes \\
teacher-forced (no sampling) & --- & $0.979$ & $0.967$ & $0.964$ & yes \\
\bottomrule
\end{tabular}
\end{table}

The ordering survives all three. The roots also differ because these measurements use a
separate prompt pool and baseline, and the token random-number generator is not reset between
them. A measurement with each arm's own sampler was also taken, but it lacks a matched
generation-0 baseline and supports no crossing comparison. Teacher-forced averages are also
distinct from document entropy: $\E_t[H_{1,t}]$ and $\E_t[2^{H_{1,t}}]$ need not move together.

\paragraph{Concentration and thresholded support.} At training $T=1$, the teacher-forced mean
$H_1$ rises by $0.75$ bits and $H_2$ by $0.84$, and the mass outside the top 50 tokens rises from
$0.226$ to $0.286$. Meanwhile the mean number of tokens with probability above $10^{-6}$ falls
from $6{,}728$ to $3{,}665$ (46\%), and the mean effective vocabulary $\E_t[2^{H_{1,t}}]$ falls
by 25 tokens. The mathematical support of the dense softmax stays fixed. At training $T=1.5$,
$H_2$ rises by $10.8$ bits, the tail mass by $0.76$, and the thresholded support gains about
$43{,}000$ tokens, indicating much flatter predictions.

\subsection{Three training decoders}
\label{sec:threepoint}

To test stronger over-dispersion than the untruncated arms provide, we add a third arm at
training $T=1.5$ and top-$p=0.95$, with three seeds and nine generations, in Spanish,
Qwen-Instruct and Chinese (Table~\ref{tab:threepoint}). English GPT-2 and Hindi gpt2-hindi have
an analogous arm at training $T=2$ and top-$p=0.95$, also with three seeds and nine generations.
Both enter Appendix~\ref{sec:vocabdrift}, and the English arm is also part of the detection pool.

At generation~0, before any self-training, sampling at $T=1.5$ raises empirical $\Htwo$
relative to the measurement sampler: $11.53$ against $6.32$ bits in Spanish, $11.03$ against
$7.62$ for the instruction-tuned model, and $9.13$ against $8.20$ in Chinese. These
document-level observations agree in direction with Proposition~\ref{thm:temp} but do not follow from
it, since context variation and the sampling path also affect document counts.

\begin{table}[ht]
\centering\small
\caption{Generation-9 $\Htwo$ under three training samplers, three-seed means of the logged two-batch
measurement samples. Their budgets are unequal: the matching eight-batch generation-9 dumps
shrink to 10.4--11.1k words for the instruction-tuned $T=1.5$ arm and grow to about 41k for the
Spanish one, and the logged samples are about a quarter as long, so the ordering is descriptive
and unequal-budget bias is not controlled. Spanish spans $8.01$ bits. With the
penalty, the instruction-tuned setting keeps the arm ordering but places every arm above
generation~0; without it, the truncated arm falls below generation~0.
$^\ast$Sampled with the checkpoint's default repetition penalty of 1.1. $^{\dagger}$About 145 Han
characters per corpus: the entropy is defined, but coverage is insufficient for this
comparison.}
\label{tab:threepoint}
\begin{tabular}{lcccc}
\toprule
Setting & gen.~0 & truncated & untruncated & hot \\
 & & (top-$p$ $0.80$) & (top-$p$ $1.0$) & ($T{=}1.5$) \\
\midrule
Spanish full-FT, GPT-2 & $6.32$ & $4.80$ & $6.39$ & $12.81$ \\
English LoRA, Qwen-Instruct$^\ast$ & $7.62$ & $8.13$ & $9.82$ & $11.32$ \\
English LoRA, Qwen-Instruct, no penalty & $6.57$ & $5.54$ & $6.68$ & $11.31$ \\
\textbf{Chinese} full-FT, Wenzhong & $8.20$ & $6.28$ & $7.02$ & n/a$^{\dagger}$ \\
\bottomrule
\end{tabular}
\end{table}

In Spanish, $\Htwo$ increases from the truncated to the untruncated to the hot arm across an
$8.01$-bit span, and at a fixed 13,000 words the hot arm's generation-9 text has 10,243 word
types, against 4,095 at generation~0. The instruction-tuned checkpoint keeps the same arm ordering,
$8.13<9.82<11.32$, although all three values exceed generation~0; without the repetition penalty
the ordering is $5.54<6.68<11.31$ and only the truncated arm falls below generation~0.

\paragraph{Chinese script loss.} High-temperature Chinese training nearly eliminates Han
script. Han characters fall from 83.8\% of all emitted characters (including whitespace) at generation~0 to 82.1\% at
generation~1, then 4.8\% at generation~2, 0.7\% at generation~3 and 0.05\% by generation~9, an
average of 145 Han characters per sampled corpus spread over 36--125 distinct types. The model
emits fragmented Latin-script text instead (``\emph{Authorization490 didn entityraceiddler time
Exploset Change conclusionThus}'').

On the eight-batch dumps these Han characters give $\Htwo=4.78$ bits ($4.47$ on the logged
two-batch sample). With at most 125 observed types the value cannot exceed $\log_2 125\approx7$
bits and is not comparable to a 13,000-character estimate, so we exclude this cell from the entropy comparison and report script coverage instead: the
restricted unit discards almost all of the output while still returning a finite score.
Tokenizer-level cross-perplexity registers the damage ($18.3\times$).

\subsection{Tokenizer composition and script drift}
\label{sec:vocabdrift}

At infinite decoding temperature a fixed softmax with finite logits tends to the uniform
distribution over its vocabulary. This suggests a simple diagnostic: decode each vocabulary entry
and measure its character composition. It is only a heuristic for finite-temperature recursive
training, which need not converge to a uniform law, and byte-level tokenizers need care because
decoding a token on its own can differ from decoding it within a sequence.

\begin{table}[ht]
\centering\small
\caption{Tokenizer composition and observed script share under high-temperature self-training.
``Bearing tokens'' counts entries of each tokenizer's $50{,}257$-entry base vocabulary, whose separate decoding contains
a character of the target script; ``vocabulary \%'' is that script's share of characters in the
decoded vocabulary. Observed shares are of all characters, including whitespace, at training $T=2$
(English, Hindi) or $T=1.5$ (Chinese); they are three-seed means, and errors are percentage
points rounded to one decimal. Vocabulary composition serves only as a descriptive reference.}
\label{tab:vocabdrift}
\begin{tabular}{llccccc}
\toprule
 & & \multicolumn{2}{c}{tokenizer} & \multicolumn{2}{c}{observed} & \\
\cmidrule(lr){3-4}\cmidrule(lr){5-6}
Setting & script & bearing tokens & \textbf{vocabulary \%} & gen.~0 & \textbf{gen.~9} & error \\
\midrule
English, GPT-2 & a--z & $44{,}170$ & $79.5$ & $75.7$ & $\mathbf{79.3}$ & $0.2$ \\
Hindi, gpt2-hindi & Devanagari & $19{,}171$ & $28.0$ & $77.1$ & $\mathbf{23.9}$ & $4.1$ \\
Chinese, Wenzhong & Han & $51$ & $0.0$ & $83.8$ & $\mathbf{0.05}$ & $0.0$ \\
\bottomrule
\end{tabular}
\end{table}

The observed endpoints lie close to this reference in English and Chinese and differ by $4.1$
percentage points in Hindi (Table~\ref{tab:vocabdrift}). Jointly decoding uniformly sampled token
streams gives similar reference shares, so the large
Chinese mismatch with the training corpus is not just an artifact of decoding tokens one at a
time. Wenzhong represents Han characters as multi-token byte sequences, and only 51 separate
entries contain Han characters, which is consistent with its
vulnerability to script loss under diffuse sampling.

\section{Detection experiments and interventions}
\label{app:detect}

\begin{table}[ht]
\centering\small
\caption{Alarm rules evaluated over generations 1--9: number of trajectories on which each rule
ever fires. ``Above cutoff'' trajectories reach at least twice their generation-0
cross-perplexity; the no-training and human-adaptation controls are described below.}
\label{tab:alarms}
\begin{tabular}{lccc}
\toprule
rule & above cutoff & no training & human adaptation \\
\midrule
magnitude: $M_g>1$ bit & $92/95$ & $0/6$ & $0/3$ \\
profile: $M_g>1$ or $D_g$ & $95/95$ & $0/6$ & $3/3$ \\
profile with no sign floor & $95/95$ & $6/6$ & $3/3$ \\
\bottomrule
\end{tabular}
\end{table}

\paragraph{Classification protocol.} The pool comprises 660 corpora from 16 settings, 277 of
them labeled degraded because their cross-perplexity is at least twice the generation-0 value.
Features are changes from generation~0 at a budget of 12,000 words (12,000 Han characters
for Chinese), and 42 shorter corpora are excluded: 24 from the Chinese high-temperature arm and 18 from the instruction-tuned
high-temperature arm. The pool has 78 trajectories in 48 (setting, seed) blocks. Folds hold out
experimental settings, so model families can occur in both the training and the evaluation
partitions. The Chinese and Qwen-Instruct high-temperature folds contain a single class
and are excluded from AUROC means;
the Spanish high-temperature fold has just one positive among 27 rows, and the English $T=2$ fold
has 24 positives among 27. The four Qwen2.5-0.5B-Instruct settings were sampled with the
checkpoint's repetition penalty of 1.1. Runs without saved text are not included.

A single feature is scored directly, with its sign chosen by its AUROC on the training partition.
A feature pair is standardized on the training partition and fit by L2-regularized logistic
regression (inverse regularization strength 1, LBFGS, at most 2,000 iterations). The nested
procedure chooses among all 15 two-feature pairs by inner leave-one-setting-out AUROC and then
scores the chosen pair on the outer setting; the fixed-pair analysis in Table~\ref{tab:detect}
evaluates a different procedure. The pooled in-sample interval of Appendix~\ref{app:posthocanalyses} resamples
complete (setting, seed) blocks, preserving all arms and generations within a block. It does not
model dependence between settings that share a checkpoint or corpus, so it is not a generalization
interval.

\begin{table}[ht]
\centering\small
\caption{Pooled correlations and fixed-feature leave-one-setting-out AUROCs on 660 corpora from
16 settings. Each feature set is held fixed across folds, and the mean AUROC excludes the two
single-class folds; the last three rows are individual held-out settings, with $n_+$ positives. Nested selection instead chooses $(\Htwo,\text{distinct-}2)$ in 11 of 14
folds and has mean AUROC $0.897$. Shared model families limit this to
transfer across conditions. $^\ast$Sampled with the checkpoint's repetition penalty of 1.1.}
\label{tab:detect}
\setlength{\tabcolsep}{4pt}
\begin{tabular}{lcccccccc}
\toprule
 & $\Hz$ & $H_1$ & $\Htwo$ & $\Hinf$ & distinct-1 & distinct-2 & $\Hz{+}\Htwo$ & $\Htwo{+}$distinct-2 \\
\midrule
Spearman $\rho$, log cross-ppl & $-0.52$ & $-0.39$ & $-0.22$ & $-0.04$ & $-0.48$ & $-0.42$ & --- & --- \\
\midrule
mean held-out AUROC & $0.818$ & $0.779$ & $0.650$ & $0.575$ & $0.817$ & $0.860$ & $0.961$ & $0.963$ \\
\;\;EN $T{=}2.0$ ($n_{+}{=}24$) & $0.000$ & $0.000$ & $0.000$ & $0.000$ & $0.000$ & $0.340$ & $0.972$ & $1.000$ \\
\;\;Qwen-Instruct$^\ast$ ($n_{+}{=}29$) & $0.666$ & $0.606$ & $0.412$ & $0.303$ & $0.662$ & $0.648$ & $0.906$ & $0.909$ \\
\;\;ES high-$T$ ($n_{+}{=}1$) & $0.038$ & $0.038$ & $0.231$ & $0.192$ & $0.038$ & $0.423$ & $0.692$ & $0.808$ \\
\bottomrule
\end{tabular}
\end{table}

Adding the coarse grid as one more held-out condition, with its 13 cells at top-$p\ge0.6$ (585
corpora), the fixed pair $(\Hz,\Htwo)$ scores AUROC $0.926$ on it, against $0.556$ for richness
and $0.491$ for collision entropy alone. Relabeling the 660-corpus pool at $1.5$, $2$, $3$ and $5$
times the generation-0 cross-perplexity gives fixed-pair mean AUROCs of $0.966$, $0.961$, $0.966$
and $0.918$, and richness alone gives $0.870$, $0.818$, $0.861$ and $0.788$. With absolute rather
than change features, the pair gives $0.983$, richness $0.829$ and min-entropy $0.342$. These are
exploratory fixed-feature comparisons on this pool and are sensitive to the unit and to fold
composition.

\paragraph{Lead time.} We score each coarse-grid trajectory at generations 1--9 against its own
baseline and compare the generation at which an alarm first fires with the generation at which
cross-perplexity first reaches the $2\times$ threshold. All 75 trajectories eventually degrade, so
this experiment cannot estimate a false-alarm rate. The profile rule fires if either magnitude
exceeds one bit, or if the signs disagree with both magnitudes above $0.20$ bits
(Table~\ref{tab:leadtime}).

\begin{table}[ht]
\centering\small
\caption{Lead time of alarm rules on the 75 coarse-grid trajectories. Positive lead means the rule
fires before the quality threshold is reached. A trajectory counts as reversing if, at some
generation, a rule that fired at the previous generation stops firing. The distinct-2 and rep-4
rules fire when the absolute change exceeds 20 times the largest difference between generation-0
texts ($0.309$ and $0.029$). $^\dagger$Both orders must exceed the $0.20$-bit floor for the sign
clause to fire.}
\label{tab:leadtime}
\begin{tabular}{lccccc}
\toprule
rule (fires when\ldots) & fires on & median lead & range & never fires & reverses \\
\midrule
$|\Delta\Hz|>1$ or $|\Delta\Hinf|>1$ or signs disagree$^\dagger$ & $\mathbf{75/75}$ & $\mathbf{+1}$ & $-1$ to $+3$ & $\mathbf{0}$ & $\mathbf{0}$ \\
$|\Delta\Htwo|>1$ & $65/75$ & $+1$ & $-6$ to $+2$ & $10$ & $1$ \\
$\Delta\Htwo<-1$ (one-sided) & $33/75$ & $-1$ & $-6$ to $+1$ & $42$ & $1$ \\
$\Delta\Hz<-1$ (one-sided) & $40/75$ & $+1$ & $+0$ to $+1$ & $35$ & $0$ \\
distinct-2 movement & $40/75$ & $+0$ & $-2$ to $+1$ & $35$ & $0$ \\
rep-4 movement & $40/75$ & $+1$ & $+0$ to $+1$ & $35$ & $0$ \\
\bottomrule
\end{tabular}
\end{table}

The profile rule fires on all 75 trajectories, with a median lead of one generation and no
reversals. Keeping the sign floor but replacing the one-bit margins with larger margins scaled to
the generation-0 null (about $2.8$ bits for richness and $3.4$ for min-entropy) reduces this to 63 of 75 trajectories,
with a median lead of $-1$ generation (range $-4$ to $+3$), 12 that never fire and 11 reversals.

\paragraph{Healthy controls and an independent quality measure.} Six no-training trajectories
keep cross-perplexity at $1.00\times$. At a fixed 13,000 words every change in $H_0$, $H_1$ and
$H_2$ stays within $0.15$ bits, and the largest absolute min-entropy change is $0.240$ bits. The
unfloored sign rule fires on all six and the floored rule on none. Three trajectories adapted on
human text instead improve cross-perplexity to $0.60\times$, with mean $\Delta H_0=-0.636$ and
$\Delta H_\infty=+0.433$ at the same budget, and the floored rule fires on all three. These
controls use top-$k=50$, top-$p=0.95$ and $T=1$, prompts from the held-out book, and smaller
batches than the main grid, and all six no-training chains share this one sampler. The seed fixes
the prompts; the global token-sampling seed was not recorded. Human adaptation trains on
sequential 128-token chunks of the second half of \emph{Persuasion}, whose first half is the
evaluation reference, without gradient clipping and with a different optimizer budget. Under this protocol, the floored alarm can fire while reference likelihood improves; calibrating
the main grid requires matched prompts, samplers and optimizer budgets. An exploratory MAUVE calculation falls from $0.707$ to $0.079$
at $T=1$, top-$p=1$ (five seeds), and gives $0.022$ at $T=0.70$ (two seeds). The feature script
mean-pools final-layer states and defaults to GPT-2, 200 passages and 128 tokens per passage.
Retained features have width 768, inconsistent with GPT-2-large, but the invocation, passage
length limit and exact model revision were not recorded. We therefore cannot identify the featurizer with
certainty or claim exact reproducibility of these MAUVE values. Two high-temperature endpoints
have insufficient passage coverage.

\paragraph{Frozen external detector.} A serialized detector and threshold were frozen before the
Pythia evaluation and recorded by SHA-256. Obtaining a healthy stratum required trying pure
recursion, then 25--75\% real-data mixing (all nine runs degraded), then a lower learning rate as well; the six healthy
runs mix in 50\% or 75\% real data. The first,
pure-recursion stage used four samplers (top-$p=0.6$, $0.8$ and $1$ at $T=1$, and top-$p=1$ at
$T=1.25$) with five seeds each, later extended to eight. Every run degraded, and the 20 five-seed
runs form the Pythia part of the alarm-rule pool in \S\ref{sec:detect}. In the resulting sample
of six healthy and six degraded runs, the ranking was perfect, but the $0.5$ classifier threshold
misclassified three healthy trajectories and the one-bit rules misclassified five or six. Healthy
labels were confounded with the training recipe, and the design was adapted after initial
outcomes, so the regime-controlled follow-up in \S\ref{sec:detect} is the stronger test. Its arm
summaries are in Table~\ref{tab:pythia}.

\begin{table}[ht]
\centering\small
\caption{Regime-controlled Pythia-160m endpoints (five seeds per arm).}
\label{tab:pythia}
\begin{tabular}{lccc}
\toprule
arm (same training regime) & median ppl ratio & median $\Delta\Hz$ & median $\Delta\Htwo$ \\
\midrule
untruncated, $T=1$ & $1.94\times$ & $-1.21$ & $-0.26$ \\
truncated, top-$p{=}0.80$ & $6.30\times$ & $-4.75$ & $-2.38$ \\
hot, $T{=}2.0$, top-$p{=}0.95$ & $3.83\times$ & $\mathbf{+1.53}$ & $\mathbf{+5.80}$ \\
\bottomrule
\end{tabular}
\end{table}

All 15 runs use pure recursion, learning rate $10^{-5}$, one epoch and the same corpus, with
top-$k$ disabled; only temperature and top-$p$ vary. Twelve exceed the $2\times$ label threshold.
Movement AUROC is $0.944$ for both richness and the two-order maximum, with only three negative
runs, all in the untruncated arm. As \S\ref{sec:detect} notes, 30 of the 36 comparisons are across
decoders, and all of those are concordant, while only 4 of the 6 within-decoder comparisons are.
The within-decoder ranking remains uncertain. This exploratory score uses the classifier
comparison's $(H_0,H_2)$ pair; its feature choice and the earlier $(H_0,H_\infty)$ alarm rules
require separate validation.

\paragraph{Checkpoint-branch intervention.} To separate the effect of injecting human text from
backend and resume effects, we train one trunk per seed, save its state at every generation, and
branch at $b\in\{1,3,5,8\}$. Each of five seeds has an injected branch (25\% human text) and a
resumed placebo branch from the same state. A separate ten-seed comparison of five independently rerun policies found different
measurement texts before either injection in 56 of 100 within-seed policy pairs. The injected text is the second
half of the novel whose first half is evaluated, so the passages are disjoint but the book and
domain are shared. This trainer uses sequential chunks without gradient clipping. It discards
the final 256-token chunk even when complete (otherwise the partial tail); because injection
appends text to the corpus, the realized injected share is 24.1--24.3\%. Table~\ref{tab:branch} compares each branch with its
own placebo.

\begin{table}[ht]
\centering\small
\caption{Checkpoint-branch intervention: medians over five seeds. The unbranched no-injection
endpoint is $30.7\times$ its generation-0 cross-perplexity. The third row gives the same ratio at
the injection generation itself (lag 0) instead of at generation~9. The richness row compares each
branch with its placebo on the first 13,000 words of their generation-9 measurement texts.}
\label{tab:branch}
\begin{tabular}{lcccc}
\toprule
inject at generation $b$ & $1$ & $3$ & $5$ & $8$ \\
\midrule
final cross-perplexity, injected & $18.1\times$ & $14.1\times$ & $7.5\times$ & $\mathbf{2.3\times}$ \\
\;\;ratio to its own placebo & $0.63$ & $0.46$ & $0.27$ & $\mathbf{0.07}$ \\
\;\;same, at matched lag $0$ & $0.89$ & $0.58$ & $0.24$ & $0.06$ \\
seeds improved (of 5) & $4$ & $5$ & $5$ & $5$ \\
$\Hz$ recovered (bits) & $-0.07$ & $+0.23$ & $+0.30$ & $\mathbf{+0.89}$ \\
\bottomrule
\end{tabular}
\end{table}

Later injections have larger effects, both at generation~9 and at a matched lag after injection,
but model state and branch time are confounded. No injection restores the initial quality;
even the latest leaves the endpoint at $2.3$ times its generation-0 cross-perplexity. Richness
recovery also grows with the injection generation, from $-0.07$ bits at $b=1$ to $0.89$ bits at
$b=8$, so at equal length only the latest branch recovers much of the lost richness.
With five seeds an exact two-sided sign test cannot reach $p<0.05$, so we report effects and seed
counts.

A second run matches the main \emph{trainer} (shuffling, gradient clipping and full chunks), and a
third changes the injection pool to the King James Bible (Table~\ref{tab:branchpool}). Both
matched runs read NLTK word lists joined by spaces for prompts, evaluation and injected text;
the sequential run reads the raw novel. Their no-injection endpoints are $62.3\times$ and
$63.4\times$, versus $30.7\times$ for the sequential run. Comparing columns 1 and 2 therefore
confounds trainer and text rendering; each column is interpreted relative to its own placebo.
Columns 2 and 3 share a rendering. Matching the trainer also leaves prompt sources and lengths,
evaluation blocking, batch sizes and random-number schedules different from the main harness.

\begin{table}[ht]
\centering\small
\caption{Branch ratios of final cross-perplexity to the branch's own placebo (seeds improved of
5), for the sequential trainer (raw text), a trainer matched to the main harness, and the
matched trainer with a disjoint injection pool (both using space-joined NLTK word lists).
Richness rows use the first 13,000 words of generation~9. ``Ordered in $\alpha$'' means the
median recovery is non-increasing over $\alpha=0,1,2,\infty$.}
\label{tab:branchpool}
\begin{tabular}{lccc}
\toprule
inject at generation $b$ & sequential trainer & matched trainer & \textbf{disjoint pool} \\
\midrule
$1$ & $0.63$ \,(4/5) & $0.73$ \,(5/5) & $\mathbf{1.15}$ \,(1/5) \\
$3$ & $0.46$ \,(5/5) & $0.41$ \,(5/5) & $\mathbf{1.10}$ \,(2/5) \\
$5$ & $0.27$ \,(5/5) & $0.17$ \,(5/5) & $0.55$ \,(5/5) \\
$8$ & $0.07$ \,(5/5) & $0.03$ \,(5/5) & $0.19$ \,(5/5) \\
\midrule
$\Hz$ recovered at $b{=}8$ (bits) & $+0.89$ & $+0.92$ & $+0.21$ \\
ordered in $\alpha$ at $b{=}8$? & yes & yes & \textbf{no} \\
\bottomrule
\end{tabular}
\end{table}

Early Bible injection worsens the median endpoint ratio, although one of five seeds improves at
$b=1$ and two at $b=3$. Late injection helps all five, but its residual cross-perplexity relative
to placebo is $0.19$, against $0.03$ for the same-book matched-trainer run, about seven times as large using the unrounded ratios. Richness recovery at $b=8$ falls
from $0.92$ bits with the same book to $0.21$ bits with the Bible, under a quarter of the
same-book recovery, and the ordering across $\alpha$ is lost. The placebo's generation-9 cross-perplexity is within 5\% of the trunk's in 12 of 20 sequential-trainer
and 6 of 20 matched-trainer cases (median relative differences of $0.011$ and $0.11$); the matched-trainer run with the Bible pool gives 7 of 20
(median $0.079$). In no case are the two identical. This backend divergence is why each branch is compared with its
own resumed placebo.

\section{Additional held-out works}
\label{app:multiwork}

We reevaluate 100 GPT-2 grid endpoints and 15 Pythia endpoints on three further NLTK Gutenberg
works: Austen's \emph{Sense and Sensibility}, Melville's \emph{Moby Dick} and Burgess's \emph{The
Adventures of Buster Bear}. These works were selected before evaluation and excluded from the
five-book training corpus; their exposure during pretraining is unknown. Each contributes 60
equally spaced, disjoint 256-token blocks (15,300 predicted tokens), and the training-time
evaluation on \emph{Persuasion} uses its first 60 blocks. The CPU evaluation sums shifted-logit
losses in float64, independently of the training loss helper, and reproduces the perplexities
recorded during training within a relative error of $1.5\times10^{-5}$. Token hashes and block losses are
released. Ratios use each model family's pretrained baseline on the same work
(Table~\ref{tab:multiwork}).

\begin{table}[ht]
\centering\small
\caption{Dependence of the quality label on the reference text, at generation~9. GPT-2 values
are mean cross-perplexity ratios over five seeds at top-$p=1$, $T=1$. Pythia keeps the same
$\max(|\Delta\Hz|,|\Delta\Htwo|)$ score and $2\times$ cutoff. A dash marks a single label class.
The final row first averages log ratios over the three additional works within each endpoint; no
score or threshold is refitted.}
\label{tab:multiwork}
\begin{tabular}{lrrr}
\toprule
reference & GPT-2 $T=1$ & Pythia above $2\times$ & Pythia AUROC \\
\midrule
\emph{Persuasion} & $3.83\times$ & $12/15$ & $0.944$ \\
\emph{Sense and Sensibility} & $3.43\times$ & $11/15$ & $0.977$ \\
\emph{Moby Dick} & $4.48\times$ & $15/15$ & --- \\
\emph{The Adventures of Buster Bear} & $4.47\times$ & $15/15$ & --- \\
three additional works, geometric mean & $4.10\times$ & $15/15$ & --- \\
\bottomrule
\end{tabular}
\end{table}

On \emph{Sense and Sensibility}, Pythia's 43 of 44 concordant comparisons consist of 40 of 40
across decoders and 3 of 4 within the untruncated arm. The other additional references, and their
aggregate, leave no negative class.

\section{Proofs}
\label{app:proofs}

Throughout, $C(p)=\sum_i p_i^2$ and $\Htwo=-\log_2 C$; logarithms are base 2 unless written $\ln$.

\subsection{Tempering, sampling and the scope of the crossing illustration}
\label{app:crossings}

These are the statements behind \S\ref{sec:crossings}. Fix a distribution $q$ on a finite alphabet with $q_i>0$, and write
\[
\psi(\beta)\;=\;\ln\sum_i q_i^{\beta},
\qquad\text{so}\qquad \psi(1)=0,\qquad H_\alpha(q)=\frac{\psi(\alpha)}{(1-\alpha)\ln2}\ \text{ bits}.
\]
$\psi$ is convex, being a log-sum-exp of the linear functions $\beta\mapsto\beta\ln q_i$. Let $q^{(\beta)}_i=q_i^\beta/\sum_j q_j^\beta$ denote the escort distribution at $\beta$, and recall the standard identities $\psi'(\beta)=\mathbb{E}_{q^{(\beta)}}[\ln q]$ and $\psi''(\beta)=\operatorname{Var}_{q^{(\beta)}}(\ln q)\ge0$.

\begin{proposition}[Tempering acts by dilation]
\label{prop:temper}
For every finite $\alpha\ne1$ and $\beta>0$,
\[
H_\alpha\!\left(q^{(\beta)}\right)\;=\;\frac{\psi(\alpha\beta)-\alpha\,\psi(\beta)}{(1-\alpha)\ln2},
\qquad\text{and}\qquad
H_1\!\left(q^{(\beta)}\right)\;=\;[\psi(\beta)-\beta\,\psi'(\beta)]/\ln2.
\]
\end{proposition}
\begin{proof}
$\sum_i\big(q_i^\beta/\sum_j q_j^\beta\big)^\alpha=\big(\sum_i q_i^{\alpha\beta}\big)\big/\big(\sum_j q_j^{\beta}\big)^{\alpha}$. Taking $\ln$ gives $\psi(\alpha\beta)-\alpha\psi(\beta)$, and dividing by $(1-\alpha)\ln2$ gives the first claim; $\beta=1$ recovers $H_\alpha(q)$ since $\psi(1)=0$. For the second, apply L'H\^opital's rule in $\alpha$ at $\alpha=1$: the numerator's derivative is $\beta\psi'(\alpha\beta)-\psi(\beta)$ and the denominator's is $-\ln2$. At $\beta=1$ this is $-\psi'(1)/\ln2=-\sum_i q_i\log_2 q_i$, the Shannon entropy, as required.
\end{proof}

\begin{proposition}[Temperature sensitivity]
\label{prop:sens}
Let $G(\alpha)=-\partial_\beta H_\alpha(q^{(\beta)})\big|_{\beta=1}$, the rate at which sharpening reduces $H_\alpha$. Then $G(\alpha)\ge0$ for all $\alpha\ge0$, with
\[
G(\alpha)=\frac{\alpha}{\ln2}\frac{\psi'(\alpha)-\psi'(1)}{\alpha-1}
=\frac{\alpha}{\ln2}\int_0^1\psi''(1+s(\alpha-1))\,ds,
\]
with $G(1)=\operatorname{Var}_q(\ln q)/\ln2$ by continuity, and $G(0)=0$; as $\alpha\to\infty$, $G(\alpha)\to H_1(q)-H_\infty(q)\ge0$.
\end{proposition}
\begin{proof}
Differentiating Proposition~\ref{prop:temper} gives
$\partial_\beta H_\alpha=\alpha[\psi'(\alpha\beta)-\psi'(\beta)]/((1-\alpha)\ln2)$.
The divided difference of $\psi'$ equals the displayed integral. Since
$\psi''=\Var_{q^{(\beta)}}(\ln q)\ge0$, $G\ge0$. At order zero, tempering preserves the set of
positive probabilities, so population $H_0$ is constant. Finally, $\psi'(\alpha)\to\ln\max_i q_i$ as
$\alpha\to\infty$; substituting in the divided difference gives the stated limit.
\end{proof}

\emph{Recursive tempering.} For non-uniform $q$, $\psi''>0$, so the derivative above is negative at every $\beta>0$ for finite $\alpha>0$, $\alpha\ne1$. The limiting orders satisfy $\partial_\beta H_1=-\beta\psi''(\beta)/\ln2<0$ and $\partial_\beta H_\infty=[\psi'(\beta)-\ln\max_i q_i]/\ln2<0$. Since $(q^{(\beta)})^{(\beta')}=q^{(\beta\beta')}$, $g$ untruncated infinite-sample steps at fixed temperature $T$ give $q^{(T^{-g})}$ for every integer $g\ge1$. Relative to $q$, every positive-order entropy rises for $T>1$, falls for $0<T<1$ and is unchanged at $T=1$, while $H_0$ stays fixed.

\begin{proposition}[Collision bias and expected occupancy]
\label{prop:rarefy}
For $N\ge2$ i.i.d.\ draws from $q$, with counts $n_i$, empirical distribution $\hat q$ and $\widehat C=C(\hat q)$,
\[
\E\hat C=C+\frac{1-C}{N},\qquad
\E\!\left[\frac{\sum_i n_i(n_i-1)}{N(N-1)}\right]=C,
\qquad C=\sum_i q_i^2.
\]
Consequently $H_2(q)-\E H_2(\hat q)\le\log_2(1+(1-C)/(NC))$ by Jensen's inequality, which leaves
the sign of the deficit undetermined. For the observed support $S_N$,
$\E S_N=\sum_i[1-(1-q_i)^N]$ \citep[Eq.~14]{hurlbert1971}.

For a triangular sequence $q^{(N)}$ on $K_N\to\infty$ types (an alphabet size, unrelated to
Yule's $K$), suppose a fraction $\phi\in(0,1]$ of the types satisfy $c_1/N\le q_i\le c_2/N$ with
$0<c_1\le c_2<\infty$, and the remaining types (if any) satisfy
$\min_{i\,\mathrm{heavy}}Nq_i/\ln K_N\to\infty$. Then
\[
\phi K_Ne^{-c_2}(1+o(1))\le\E(K_N-S_N)\le \phi K_Ne^{-c_1}(1+o(1)).
\]
When $c_1=c_2=c$, $\log_2 K_N-\log_2\E S_N\to-\log_2(1-\phi e^{-c})$. The expected number of missing types
is $\Theta(K_N)$, whereas this logarithmic deficit is $\Theta(1)$.
\end{proposition}
\begin{proof}
Binomial marginals give $\E n_i^2=Nq_i(1-q_i)+N^2q_i^2$ and
$\E[n_i(n_i-1)]=N(N-1)q_i^2$, which prove both collision identities.
Convexity of $-\log_2$ gives the bound, and summing occupancy indicators gives $\E S_N$.
For light types, $(1-q_i)^N=e^{-Nq_i}(1+O(Nq_i^2))$ uniformly, so their probabilities of being
missing lie between $e^{-c_2}$ and $e^{-c_1}$ up to a factor $1+o(1)$. Each heavy type is absent
with probability at most $e^{-Nq_i}=K_N^{-\omega(1)}$, so the heavy types contribute $o(1)$ in
total. Summing proves the bound, and with $c_1=c_2=c$,
$\E S_N=K_N(1-\phi e^{-c}+o(1))$, which proves the limit.
\end{proof}

The logarithmic deficit above differs from the plug-in deficit $\log_2 K_N-\E\log_2 S_N$, which
is at least as large by Jensen's inequality. Under the proposition's assumptions the difference tends to zero.
Writing $\mu_N=\E S_N=\Theta(K_N)$, occupancy indicators have nonpositive pairwise covariance, so $\Var
S_N\le\mu_N$. Chebyshev's inequality gives $\Pr(|S_N-\mu_N|>\mu_N/2)=O(K_N^{-1})$. On the complementary event,
$|\ln(S_N/\mu_N)|\le2|S_N/\mu_N-1|$ has expectation $O(K_N^{-1/2})$; on the remaining event, $1\le
S_N,\mu_N\le K_N$ bounds its contribution by $O(\ln K_N/K_N)$. Thus the two richness deficits have the same
limit when $c_1=c_2$. Appendix~\ref{app:rarefy-fact} computes both. The lower mass bound is essential: with $K_N=N^2$
and $q_i=N^{-2}$, $\E S_N=\Theta(N)$ and the logarithmic deficit grows as $(1+o(1))\log_2 N$.
Population support cannot be recovered exactly from a finite sample without further assumptions
about unseen types \citep{good1953,chao1984}. When $\phi<1$, the heavy-type condition forces
$K_N\ln K_N/N\to0$, so the total mass of light types is $O(K_N/N)\to0$; the proposition
does not describe a tail of fixed positive mass in that case.

The logarithm introduces bias even when the collision-probability estimator is unbiased. For
$q=(1/2,1/2)$ and $N=2$, the plug-in entropy deficit is $0.5$ bits while its Jensen bound is
$\log_2(1.5)=0.585$, and the U-statistic can be zero, which makes its logarithm infinite. On a
measured English generation-0 law (the seed-0 baseline of the main grid, $\Htwo=6.96$ bits) at
$N=13{,}000$, the bound is $0.0137$ bits, the delta-method deficit is $0.0131$, and a
4,000-draw Monte Carlo estimate is $0.0128\pm0.0006$ bits (standard error). The logarithm of the U-statistic has a
residual bias estimated at $+0.0008\pm0.0006$ bits (standard error). For this non-degenerate law, $N$ exceeds
the support size, so the U-statistic is positive and nonconstant; strict Jensen inequality
makes its entropy bias strictly positive, despite Monte Carlo uncertainty. The
richness deficit on the same law is $1.69$ bits.

\paragraph{Limits of a first-order crossing explanation.} Suppose each generation lost a fixed
amount $L_\alpha$ of order-$\alpha$ entropy through finite sampling and gained $G(\alpha)(1-1/T)$
through tempering, the first-order expansion of Proposition~\ref{prop:sens}. The balance
$1-1/T^\ast=L_\alpha/G(\alpha)$ then puts any crossing above one whenever both loss and
sensitivity are positive, so it cannot produce the measured roots below one for $H_1$, whose
i.i.d.\ plug-in loss is always positive, or for $\Htwo$, whose loss is positive for large $N$
(Appendix~\ref{app:pluginsign}). A root below one requires a negative loss, which can occur at
higher orders: writing $\hat H_\alpha=H_\alpha(\hat q)$ for the plug-in estimate, exact binomial averaging for $q=(0.9,0.1)$ and $N=100$ gives
$\E\hat H_2-H_2=-0.0007285$, $\E\hat H_3-H_3=+0.0008191$ and $\E\hat H_\infty-H_\infty=+0.0008076$
bits, so orders 3 and $\infty$ gain entropy from sampling alone. Such a gain could place
$T^\ast(\Hinf)$ below one, but endpoint sensitivities do not establish a monotone ratio of loss
to gain for the intermediate orders. The
neural result is therefore empirical, and the categorical simulation provides one example of the
same outer ordering.

\paragraph{A categorical simulation.} We simulate a categorical model that alternates tempering,
13,000 draws and refitting
for nine generations. It pads the alphabet to 50,257 types, mixes each refit with a uniform
smoothing term of weight $0.02$, and measures with 13,000 further draws from a unigram analogue of the neural grid's top-$p=0.95$ measurement
sampler. Across 16 replicates the roots are $0.993$, $0.994$, $1.005$ and $1.129$ for orders
$\infty$, 2, 1 and 0. The simulation reproduces the outer ordering, with richness turning well
after the other orders, but the $\Hinf$ and $\Htwo$ roots ($0.9934$ and $0.9937$) lie in the same
grid bracket, $[0.95,1.00]$, so their order is not resolved. It does not predict the neural
roots, and without smoothing its richness does not recover in the tested range. That need not
hold in general, since a fixed nucleus sampler can expose previously excluded types even though the fitted support cannot grow (see the two-type example below). In detail, the generation-0 law is
the unsmoothed unigram law of the English generation-0 text used above; each generation tempers the current smoothed law, draws 13,000
tokens, and refits by maximum likelihood before mixing in the uniform term; and the measurement
sampler applies top-$p=0.95$ to the current smoothed law. The 14 temperatures run from $0.70$ to $2.00$
($0.70$, $0.80$, $0.90$, $0.95$, $1.00$, $1.02$, $1.05$, $1.10$, $1.20$, $1.30$, $1.45$, $1.60$,
$1.80$ and $2.00$), and replicate $i$ uses the same random seed at every temperature.

Population support and the support seen by a fixed sampler must also be distinguished. Take $q=(0.9999,0.0001)$, untruncated
training at $T=2$, $N=13{,}000$ and unsmoothed categorical refitting. A fixed top-$p=0.95$ sampler initially keeps
only the first type. Tempering can raise the second type's mass until it enters the nucleus, so
observed $H_0$ can increase without adding any type outside the original population support. A
nine-generation simulation does so in all 100 tested seeds. Thus $G(0)=0$ does not imply that
richness under a fixed sampler cannot recover under maximum-likelihood refitting.

\subsection{Unbiased estimators for integer-order power sums}
\label{app:factorial-estimator}
For integers $N\ge m\ge2$, write $(a)_m=a(a-1)\cdots(a-m+1)$. Under multinomial sampling, an
unbiased estimator of $P_m(q)=\sum_i q_i^m$ is
\[
\widehat P_{m,\mathrm u}=\frac{\sum_i(n_i)_m}{(N)_m},
\qquad \E\widehat P_{m,\mathrm u}=P_m(q).
\]
Indeed, $(n_i)_m$ counts ordered tuples of $m$ distinct sample positions that all contain type
$i$. Each tuple has probability $q_i^m$, so $\E(n_i)_m=(N)_m q_i^m$, and summing over types
proves the identity. It is the fixed-sample counterpart of the bias-corrected estimator of
\citet{acharya2017renyi}: for Poisson sample size $M\sim\mathrm{Poisson}(N)$, their numerator
$\sum_i(n_i)_m$ is divided by $N^m$, the $m$th power of the expected sample size. At order 2
the fixed-sample estimator above is the pair-count estimator. Applying the nonlinear entropy transform
$\log_2\widehat P_{m,\mathrm u}/(1-m)$ requires a separate bias analysis, and a zero estimate
gives infinite entropy.

\subsection{Order monotonicity \texorpdfstring{$H_0\ge H_1\ge \Htwo\ge H_\infty$}{H0 >= H1 >= H2 >= H infinity}}
Let $P_\alpha=\sum_i p_i^\alpha$ and $\pi_i=p_i^\alpha/P_\alpha$ (the escort distribution). In
bits,
$\frac{d H_\alpha}{d\alpha}=\frac{1}{(1-\alpha)^2\ln2}\big[\ln P_\alpha+(1-\alpha)\tfrac{P_\alpha'}{P_\alpha}\big]$
with $P_\alpha'=\sum_i p_i^\alpha\ln p_i$. Substituting $\ln P_\alpha=\alpha\ln p_i-\ln\pi_i$
reduces the bracket to $\sum_i\pi_i\ln(p_i/\pi_i)=-\ln(2)\,\KL(\pi\Vert p)\le0$. The continuous limits
cover orders $1$ and $\infty$, and zero-probability entries are omitted from the support. Hence
$H_\alpha$ is non-increasing in $\alpha$. $\square$

\subsection{Proposition~\ref{thm:temp} (temperature monotonicity)}
With $p_i=e^{\beta z_i}/Z$, $Z=\sum_j e^{\beta z_j}$ and $\mu=\sum_j p_j z_j$, the quotient rule gives $\frac{dp_i}{d\beta}=p_i(z_i-\mu)$, so
$\frac{d}{d\beta}\sum_i p_i^2=2\sum_i p_i^2(z_i-\mu)=2\Cov(p_I,z_I)$, where $I\sim p$. Taking an independent copy $I'\sim p$,
\[
2\Cov(p_I,z_I)=\E\big[(p_I-p_{I'})(z_I-z_{I'})\big].
\]
Since $p_i$ is strictly increasing in $z_i$ for $\beta>0$, the two factors share a sign, so the product is nonnegative pointwise and its expectation is nonnegative, and strictly positive unless $z$ is constant. Thus $\sum_i p_i^2$ strictly increases in $\beta$ and strictly decreases in $T$, and $\Htwo$ strictly increases in $T$. The boundary limits follow by factoring out $e^{\beta z_{\max}}$ (so $\sum_i p_i^2\to1/m$ and $\Htwo\to\log_2 m$ as $\beta\to\infty$) and from $p_i\to1/|\V|$ as $\beta\to0$, which requires finite logits ($\Htwo\to\log_2|\V|$). $\square$

\subsection{Proposition~\ref{thm:trunc} and Lemma~\ref{lem:upperset}}
\begin{lemma}[Upper-set conditioning]
\label{lem:upperset}
With $X=p_I$, $I\sim p$, and $S$ an upper set of mass $0<a<1$, $\E[X\mid I\in S]\ge\E[X]$; equality holds precisely when every retained $p_i$ and every strictly positive discarded $p_j$ share a common value.
\end{lemma}
\emph{Proof.} Let $A=\sum_{i\in S}p_i^2$, $a=\sum_{i\in S}p_i$, and let $B,b$ be the corresponding sums over the complement. Then $\E[X\mid I\in S]=A/a$, $\E[X\mid I\notin S]=B/b$ and $\E[X]=A+B$. Because $S$ is an upper set, $A/a\ge\min_{i\in S}p_i\ge\max_{j\notin S}p_j\ge B/b$. By total expectation, $\E[X]=a\,\E[X\mid I\in S]+b\,\E[X\mid I\notin S]$, and subtracting gives $\E[X\mid I\in S]-\E[X]=b\big(\E[X\mid I\in S]-\E[X\mid I\notin S]\big)\ge0$. Equality forces equality
throughout this chain. Because the weights $p_i/a$ are positive on $S$ and the weights $p_j/b$ are positive
exactly on the strictly positive discarded entries, $A/a=\min_{i\in S}p_i$ makes all retained $p_i$ equal,
$\max_{j\notin S}p_j=B/b$ makes all strictly positive discarded $p_j$ equal, and the middle link makes the two
values coincide; the converse is immediate. $\square$

\emph{Proposition~\ref{thm:trunc}.} With $q_i=p_i/a$ on $S$,
\[
\sum_{i\in S}q_i^2=\tfrac{1}{a^2}\sum_{i\in S}p_i^2=\tfrac1a\E[X\mid I\in S]\ge\tfrac1a\E[X]>\E[X]=\sum_i p_i^2.
\]
Lemma~\ref{lem:upperset} gives the first, weak inequality, and the factor $1/a>1$ makes the second strict. Applying $-\log_2(\cdot)$ reverses the inequality, so $\Htwo(q)<\Htwo(p)$. Without the upper-set condition the result fails: keeping the two smaller types of $p=(0.9,0.05,0.05)$ raises $\Htwo$ from $0.295$ to $1$ bit. $\square$

\subsection{Proposition~\ref{prop:cap} (support cap)}
For $q$ supported on $S$ with $|S|=k$, the Cauchy--Schwarz inequality gives $1=(\sum_{i\in S}q_i)^2\le k\sum_{i\in S}q_i^2$, so $\sum_{i\in S}q_i^2\ge1/k$ and $\Htwo(q)\le\log_2 k$, with equality if and only if $q$ is uniform on $S$. No reshaping of mass within $S$, at any temperature, can exceed $\log_2 k$. $\square$

\subsection{Lemma~\ref{lem:vbridge} (the \texorpdfstring{$V$}{V}-bridge)}
For a uniformly chosen position $t$, $\Var_t[p_{c_t}(v)]=\overline{p_{c_t}(v)^2}-\pbar(v)^2$.
Summing over $v$ gives $C(\pbar)=\Cbar-V$ exactly, with $V\ge0$. $\square$

Equivalently, $1-C(\pbar)=(1-\Cbar)+V$: pooled Gini--Simpson diversity is the sum
of its mean within-context value and the between-context variance \citep{rao1982,lande1996}.
This additive partition differs from a decomposition into independent diversity components
in the sense of \citet{jost2007}.

For the fixed-length path statement, let $X_t$ be the one-hot sampled token, $\mathcal F_{t-1}$
the history up to position $t-1$, and $p_t=\E[X_t\mid\mathcal F_{t-1}]$ its sampling law. The
differences $X_t-p_t$ are martingale differences, so their cross inner products have zero
expectation. Hence
\[
\E\left\|\ph_N-\pbar_N\right\|_2^2
=\frac{1}{N^2}\sum_{t=1}^N\E[1-\|p_t\|_2^2]\le\frac1N.
\]
This requires the conditionals of the actual sampler, including any truncation; softmaxes on
unrelated human contexts do not qualify. The bound alone gives neither an i.i.d.\ entropy-bias
formula for the path nor a fixed limiting distribution.

\subsection{Plug-in bias, with its sign}\label{app:pluginsign} For $N$ i.i.d.\ draws, $f\sim\mathrm{Multinomial}(N,p)$ with $\mathrm{Binomial}(N,p_v)$ marginals, $\E[\ph_v^2]=p_v^2+p_v(1-p_v)/N$, so $\E[\widehat C]=C+(1-C)/N$. For independent but non-identical conditionals the same calculation gives the $V$-bridge form $\E\,C(\ph)=C(\pbar)+(1-\Cbar)/N$. For comparison, \citet{miller1955} and \citet{paninski2003} study bias in Shannon-entropy estimation. To determine the sign of the induced bias in $\Htwo$, write $P_3=\sum_v p_v^3$. A delta-method expansion of $\hat\Htwo=-\log_2\widehat C$, using $\Var\,\widehat C=\tfrac4N(P_3-C^2)+O(N^{-2})$, gives
\[
\E\,\hat\Htwo-\Htwo=-\frac{C+C^2-2P_3}{N\,C^2\ln 2}+O(N^{-2}).
\]
The remainder is controlled because $1/|\operatorname{supp}p|\le\widehat C\le1$, so $-\log_2$ and
its derivatives are bounded on the range of $\widehat C$, and the moments about $C$,
$\E(\widehat C-C)^j$ are $O(N^{-2})$ for $j=3,4$; a fourth-order Taylor expansion with Lagrange
remainder then gives the stated error, with constants that depend on $p$. The numerator is strictly
positive for non-degenerate $p$: from $P_3=\sum_v p_v\,p_v^2\le p_{\max}C$ and $C\ge p_{\max}^2$,
\[
C+C^2-2P_3\ \ge\ C\big(1+C-2p_{\max}\big)\ \ge\ C\big(1-p_{\max}\big)^2\ >\ 0
\]
whenever $p_{\max}<1$. Hence, for all sufficiently large $N$, $\E\,\hat\Htwo<\Htwo$: the
collision-rate bias dominates the competing Jensen term, and the plug-in $\Htwo$ is biased
\emph{downward}, which supports the asymptotic entropy form of the ratchet in
Proposition~\ref{prop:ratchet}. $\square$

\subsection{Proposition~\ref{prop:ratchet} (diversity ratchet)}
\begin{proposition}[Truncated categorical refitting is a diversity ratchet]
\label{prop:ratchet}
Idealize generation $g{+}1$ as the categorical maximum-likelihood estimate (the empirical distribution) $\ph_g$ of $N$ draws from the top-$p$ truncation $q_g$ ($a<1$) of generation $g$'s law $p_g$. Then $\E\big[2^{-\Htwo(\ph_g)}\mid p_g\big]>2^{-\Htwo(p_g)}$ for every $N$.
\end{proposition}
\emph{Proof.} Truncation gives $C(q_g)>C(p_g)$ (Proposition~\ref{thm:trunc}), and
categorical refitting gives
\[
\E[C(\ph_g)\mid q_g]=C(q_g)+\frac{1-C(q_g)}{N}\ge C(q_g)>C(p_g).
\]
This proves the collision inequality for every $N$. For the entropy form, fix $p_g$ and its
retained set. The expansion above gives $\E[\Htwo(\ph_g)\mid q_g]=\Htwo(q_g)+O(N^{-1})$, and since
$\Htwo(q_g)<\Htwo(p_g)$, the expected entropy also decreases for sufficiently large $N$. The
entropy conclusion is asymptotic, whereas the collision inequality is exact. $\square$

Without truncation ($a=1$, $q_g=p_g$) the collision inequality still holds, since
$\E[C(\ph_g)\mid p_g]=C(p_g)+(1-C(p_g))/N>C(p_g)$ for non-degenerate $p_g$, and the sign result of
Appendix~\ref{app:pluginsign} shows that expected $\Htwo$ also falls for large $N$, by $O(N^{-1})$
rather than by the fixed amount that truncation removes. The rise of the neural untruncated arms
(\S\ref{sec:robust}) lies outside this model.

\subsection{Entropy and reference cross-perplexity}
For one categorical law, $2^{H_1(p)}\ge2^{H_2(p)}$ by order monotonicity. Cross-perplexity on a
reference law $r$ is $2^{H(r,p)}$, where $H(r,p)=H(r)+\KL(r\Vert p)$ in bits, which is finite if
and only if $r\ll p$ on a finite alphabet. The two quantities can move in opposite directions: for
uniform $r$ on $K>1$ types and $p_\epsilon=(1-(K-1)\epsilon,\epsilon,\ldots,\epsilon)$, as
$\epsilon\downarrow0$, $H_1(p_\epsilon)\to0$ while $H(r,p_\epsilon)\to\infty$.

\section{Representation and estimator controls}
\label{app:controls}

\subsection{Changing the representation}
\label{app:repr}

The primary representation lowercases each token after tokenization and drops tokens that
contain no letter or number;
punctuation attached to a word is kept. NLTK 3.10.3 changes the $T=1$ entropy differences
by less than $0.02$ bits and the roots by less than $0.001$. To test representation sensitivity, we recomputed the coarse temperature
axis on the \emph{same stored generations} under eight representations: the released tokenizer;
the released tokenizer with dandas stripped; Unicode canonical (NFC) and compatibility (NFKC)
normalization; non-whitespace
characters; raw UTF-8 bytes, which involve no learned vocabulary; a naive whitespace split without
lowercasing or punctuation removal; and the BPE vocabulary of Qwen2.5-0.5B, a model that generated
none of this text. We compare the \emph{signs} and the \emph{crossing order}; absolute entropies
have different units across representations and are not directly comparable
(Table~\ref{tab:repr}).

\begin{table}[ht]
\centering\small
\caption{Crossing behavior across representations of the same stored generations, counting each
distinct curve once.}
\label{tab:repr}
\begin{tabular}{p{0.52\linewidth}ccc}
\toprule
 & English grid & Hindi & total \\
\midrule
distinct measured curves & $6$ & $7$ & $13$ \\
$\Delta\Htwo<0$ at the lowest and $>0$ at the highest swept temperature & $6$ & $6$ & $12/13$ \\
outer ordering $T^\ast(\Hinf)<T^\ast(\Hz)$ & $6$ & $6$ & $12/13$ \\
full ordering & $5$ & $4$ & $9/13$ \\
\bottomrule
\end{tabular}
\end{table}

The English released, danda-stripped and NFC curves coincide, as do the Hindi released and NFC
curves. Counting each identical curve once leaves 13 distinct measurements from 16
configurations; the two NFKC curves are distinct but differ from the released tokenizer's roots by
less than $10^{-4}$. The outer ordering is unresolved for Hindi under the foreign BPE vocabulary, whose
min-entropy root is not bracketed. For Hindi bytes, $\Delta\Htwo$ is
already positive at the lowest swept temperature. The full ordering has a
reversed pair for English bytes ($T^\ast(H_2)=0.932$ against $T^\ast(H_\infty)=0.944$), Hindi
bytes and danda-stripped Hindi; it is unresolved for Hindi under the foreign
BPE vocabulary, where min-entropy does not cross in the swept interval. Crossing temperatures shift by up to $0.14$ in English and
$0.08$ in Hindi across representations. These transformations of the same texts show that the
outer ordering is more stable than the full ordering, which depends on representation.

A secondary Pythia-160m check covers only $T=1.00$ and $1.25$. Word-level $\Delta H_0$ stays
negative at both ($-2.92$ and $-0.18$ bits), leaving its crossing unresolved, while character and
byte representations bracket the outer ordering. This shorter grid is released but excluded from
Table~\ref{tab:repr}.

\paragraph{Hindi punctuation.} NLTK can leave Devanagari dandas (U+0964 and U+0965) attached to
sentence-final words, and the primary tokenizer keeps them. Stripping them reduces generation-0
types by 2.0\% and moves $(T^\ast(\Htwo),T^\ast(\Hinf))$ from $(1.020,1.000)$ to
$(1.021,1.023)$, which swaps these two roots by less than the grid resolves. The $\Hz$ and $H_1$
roots agree to three decimals, and the outer ordering persists. Every other Hindi result in the paper uses the primary tokenizer.

\paragraph{Measurement boundaries.} For $T=0.9$, $1$ and $1.1$ and five seeds, we regenerated the
measurement dumps from retained checkpoints, holding the sampled token IDs fixed, and compared
plain concatenation with newline-separated continuations. All 20 baseline and endpoint dumps
reproduce the original text byte for byte. With separators, $\Delta H_0<0<\Delta H_2$ still holds
at $T=1$ in all five seeds under both common-length and fixed 13,000-word budgets, with fixed-budget
means of $(-0.401,+0.908)$ bits, and each of the four mean-curve roots on this three-temperature
axis changes by less than $8\times10^{-5}$ under either budget. This isolates measurement
formatting while holding training fixed, and the released texts reproduce the comparison without
model inference.

\paragraph{Training boundaries and generation budgets.} The default sampler removes padding IDs and
concatenates the retained sequences without an added separator. GPT-2 uses ID 50256 for both
padding and EOS. Qwen2.5-0.5B-Instruct uses padding 151643 and stop IDs 151645 and 151643, so
only the primary EOS 151645 remains. OLMo-2-0425-1B uses padding 100277 and EOS 100257 and
retains EOS in training;
the frequency and effect of this difference were not isolated.
The German and Hindi checkpoints declare EOS ID 50256, which in their tokenizers is an ordinary
subword (``riegel'' and ``\,\.{G}excav''), and their padding IDs (50265 and 50257) lie one past
the end of the embedding matrix. Their continuations therefore end early only when that subword
is sampled, after which the out-of-range padding ID is fed back for the finished row. This
was tolerated by the MPS backend used for these runs; CPU execution raises an index error. Prompts are
never padded, because every prefix has exactly 16 tokens. Early termination is rare: re-tokenizing the stored
dumps finds the subword 5 times in 38,400 German continuations and 16 times in 64,000 Hindi
continuations, all of the latter in the Hindi $T=1.25$ cell, so about 21 dumped continuations are affected. The release records generation defaults and boundary IDs for every cached sampled checkpoint; for Spanish
GPT-2, whose tokenizer files were unavailable, it records the configuration and generation defaults but not
the tokenizer-derived IDs.

The GPT-2 EOS-retaining variant truncates each continuation at its first EOS and inserts an EOS
separator before rechunking. At three seeds and common-length endpoint budgets it gives
$(\Delta H_0,\Delta H_2)$ of $(-4.12,-1.49)$ at $T=0.70$, $(-0.60,+0.62)$ at $T=1$ and
$(+1.46,+6.07)$ at $T=1.25$, preserving the low-temperature contraction, the split in the middle
and the high-temperature increase. Its generation batch is 32 rather than 64, so it changes both
boundary handling and corpus size: it shows the phenomenon with EOS retained but does not isolate
the effect of removing EOS. More generally, the cross-scale LoRA grid uses generation batch 16
and the multilingual grids use 32, with four corpus batches and eight dump batches retained, so
shifts of the interval between settings cannot be attributed solely to tuning regime or parameter
count. For the ten settings in Table~\ref{tab:robust}, generation batches are 64 for English
GPT-2 and German, 32 for Spanish, Hindi and Chinese, 24 for SmolLM2 and 16 for Qwen. LoRA uses
rank 16, $\texttt{lora\_alpha}=32$ (scaling factor 2) and no LoRA dropout; checkpoint dropout
(e.g., 0.1 for GPT-2) stays active during training. Target modules are recorded in the manifests.

\subsection{Densifying the crossing region}
\label{app:dense}

Adding $T=0.85$, $0.95$ and $1.05$ with five seeds under the batch-64 protocol changes the roots
$(H_\infty,H_2,H_1,H_0)$ from $(0.920,0.949,0.997,1.048)$ to $(0.927,0.961,0.997,1.037)$, and
adding $T=0.98$ and $1.02$ then gives $(0.927,0.962,0.997,1.035)$. The final brackets, in the same
order, are $[0.90,0.95]$, $[0.95,0.98]$, $[0.98,1.00]$ and $[1.02,1.05]$. Both of the last two
cells satisfy $\Delta H_0<0<\Delta H_\infty$ in all five seeds at common-length budgets, and at the
fixed 13,000-word budget the inequality holds for both cell means. The main text and figure use
both refinements, while the comparison with other metrics keeps the coarse grid; the released
analysis script reproduces both. A separate temperature grid with generation batch 32 halves the
training and dump budgets, so we report it only as a protocol sensitivity: combined with the
coarse cells, its common-length roots are $0.921$, $0.963$, $0.998$ and $1.041$ for orders $\infty$, 2, 1 and 0,
in the same order.

Varying the number of self-generated corpus batches (one, two or eight at generation batch 32, that is $1/8$
to $1$ times the main grid's corpus) also changes the
number of gradient steps, because the harness trains for two epochs. That experiment cannot
isolate rarefaction; the categorical experiments below vary sample size without any optimization.

\subsection{Factorial checks of the occupancy and collision formulas}
\label{app:rarefy-fact}

We test the occupancy formulas on known laws, varying sample size, support and tail shape
independently. Half of $K$ types have mass $1/N$ and the rest share the remaining mass. Choosing
$N=\lfloor K(\ln K)^2\rfloor$ makes the required ratio $\min Nq_i/\ln K$ over heavy types grow with $K$
(Table~\ref{tab:factorial}).

\begin{table}[ht]
\centering\small
\caption{Occupancy under a law with $K/2$ light types of mass $1/N$. Values are exact to the displayed precision: the
missing fraction and $\log_2(K/\E S_N)$ follow from the closed form in
Proposition~\ref{prop:rarefy}, and the plug-in deficit from an occupancy recursion over the light
types (each heavy type is absent with probability below $10^{-30}$).}
\label{tab:factorial}
\begin{tabular}{rrrrrr}
\toprule
$K$ & $N$ & $\min_{\mathrm{heavy}} Nq_i/\ln K$ & missing$/K$ & $\log_2\frac{K}{\mathbb{E}S_N}$ & $\log_2 K-\mathbb{E}\log_2 S_N$ \\
\midrule
$500$ & $19{,}310$ & $12.3$ & $0.183935$ & $0.293244$ & $0.293494$ \\
$2{,}000$ & $115{,}547$ & $15.1$ & $0.183939$ & $0.293251$ & $0.293314$ \\
$8{,}000$ & $646{,}157$ & $17.9$ & $0.183940$ & $0.293252$ & $0.293268$ \\
$32{,}000$ & $3{,}443{,}498$ & $20.7$ & $0.183940$ & $0.293252$ & $0.293256$ \\
$128{,}000$ & $17{,}701{,}447$ & $23.4$ & $0.183940$ & $0.293252$ & $0.293253$ \\
\bottomrule
\end{tabular}
\end{table}

The missing fraction is within $5\times10^{-6}$ of $\tfrac12 e^{-1}=0.18394$, and both logarithmic
deficits approach $-\log_2(1-\tfrac12e^{-1})=0.293252$ bits. The plug-in deficit exceeds
$\log_2(K/\E[S_N])$ by $2.5\times10^{-4}$ bits at $K=500$ and by $10^{-6}$ at
$K=128{,}000$.

\paragraph{Failure without the lower mass bound.} Uniform laws with $K=N^{\gamma}$, $\gamma\ge1$, have every type
of mass $O(1/N)$, but only $\gamma=1$ also gives mass $\Omega(1/N)$ (Table~\ref{tab:uniformdef}).

\begin{table}[ht]
\centering\small
\caption{Exact richness deficit $\log_2 K-\E\log_2 S_N$ for uniform laws on $K=N^\gamma$ types,
computed by the occupancy recursion over draws.}
\label{tab:uniformdef}
\begin{tabular}{rrrr}
\toprule
$N$ & $\gamma=1$ & $\gamma=2$ & $\log_2 N$ \\
\midrule
$300$ & $0.6609$ & $8.2312$ & $8.2288$ \\
$1{,}000$ & $0.6615$ & $9.9665$ & $9.9658$ \\
$3{,}000$ & $0.6616$ & $11.5510$ & $11.5507$ \\
\bottomrule
\end{tabular}
\end{table}

At $\gamma=1$ the deficit rises toward $-\log_2(1-e^{-1})=0.6617$ bits, while at $\gamma=2$ it
tracks $\log_2N$. Separately, in nine Zipf-law cells (exponents $0.8$, $1.0$ and $1.4$, 1,000
types, budgets of 4--64 times the support), the measured collision-entropy bias is negative, in
two cells by less than its Monte Carlo standard error; since the Jensen bound is a theorem, the
one cell that misses it by $0.4$ standard errors reflects Monte Carlo error.

\subsection{Finite-sample estimator checks}
\label{app:rarefy-emp}

We compare the finite-sample estimators of Appendix~\ref{app:crossings} on known distributions
\citep[cf.][]{chao2012coverage,chao2014rarefaction}.

\paragraph{Estimator comparisons.} A uniform law on 5,000 types, Zipf laws on 20,000 types ($s=0.8$, $1.0$ and $1.2$) and measured endpoint laws
are sampled at budgets from $N/8$ to $2N$, with $N=13{,}000$. On the English generation-0 law of
Appendix~\ref{app:crossings} at $N=13{,}000$, the mean pair-count collision estimate over 4,000
draws is $8.0166\times10^{-3}$ against a true value of $8.0181\times10^{-3}$. The plug-in
entropy deficit is $0.0128\pm0.0006$ bits (standard error), against the Jensen bound of $0.0137$ and the
delta-method approximation of $0.0131$, and the log U-statistic retains $+0.0008\pm0.0006$ bits (standard error); its population bias is strictly positive by the Jensen argument above. Observed
richness instead has a $1.69$-bit deficit at 31\% type coverage, and no collision correction
estimates unseen types.

Low type coverage alone does not determine the accuracy of the delta-method correction. Its
collision-rate shift relative to $C$ is $(1-C)/(NC)$, so small $NC$ can make the expansion
poor; higher moments also enter the entropy correction. On the uniform law at $N=1{,}625$
($NC=0.33$, 28\% type coverage), it predicts a $4.44$-bit deficit against a measured $2.03$,
an error of $2.41$ bits. The error falls to $0.88$ at $NC=0.65$ and $0.09$ at $NC=2.6$.
Conversely, on the Zipf law with $s=1.2$ at $N=1{,}625$, the delta-method prediction matches the Monte Carlo
deficit to within its standard error (a difference of $0.0004$ bits against a standard error of $0.0015$) at
2.4\% type coverage ($NC=94$). Coverage and the estimated collision scale $N\,2^{-\widehat H_2}$ should both be
reported; neither low coverage nor $NC$ alone is a universal accuracy criterion.

\paragraph{The reported $\Hz$ and $\Htwo$ changes are large relative to seed variability.} For each cell on the
six-temperature top-$p=1$ axis, we divide the mean common-length change from generation~0 to
generation~9 by its standard error across the five seeds, a one-sample $t$ statistic that treats
seeds as the independent units, since generations within a trajectory are not independent.
Observed changes sit $10.5$ to $442$ standard errors from zero for $\Htwo$ (median $49$) and $11.3$ to $89$ for $\Hz$ (median $51$).
These ratios quantify variability across training seeds, whereas same-law rarefaction resamples
a fixed distribution. Separate generation-0 sampling controls, compared at each pair's common
length (about 70,000 words), give a largest movement of $0.061$ bits at any of the four primary orders and a
two-point profile statistic below $0.049$. Calibrating the 13,000-word alarm rules requires controls at that smaller budget. For comparison, the $T=1$ changes in
Table~\ref{tab:crossings} are $-0.81$ bits for $H_0$ and $+0.94$ for $H_2$. An autoregressive bias correction remains a separate problem from these sampling controls and
the $0.013$-bit i.i.d.\ reference calculation.

\subsection{Post-hoc checks of order, head concentration and measurement scale}
\label{app:posthocanalyses}

The following analyses use the stored English endpoints and count multisets and, for the detection baselines,
the released 660-row feature pool; none trains a language model. They were developed after inspecting the main results. Saved outputs and analysis provenance
are released in \texttt{code/exp\_data/posthoc\_analyses/}, with scripts in
\texttt{code/scripts/posthoc\_analyses/}.

\paragraph{An interpolation-free sign statement.} We evaluate
\[
\alpha\in\{0,0.1,0.25,0.5,0.75,1,1.25,1.5,2,2.5,3,4,5,7,10,20,50,\infty\}
\]
on the eleven-temperature, top-$p=1$ English axis. At each measured temperature the changes
are either all negative, all positive, or pass once from negative to positive as order increases,
in every seed and in the mean, under both common-length and 13,000-word budgets. This is a
statement about the sampled orders; it does not assert monotonicity of the magnitudes or a
theorem for every real order. The interpolated crossing temperatures decrease strictly across
all 18 orders in 4,000/4,000 common-length bootstrap draws and 3,976/4,000 (99.4\%) fixed-budget
draws. The four primary orders retain their ordering in all 4,000 under both budgets.

\paragraph{Head decomposition.} Let $A_k$ be the $k$ most frequent types in the pooled five
generation-0 texts. For each seed we truncate generation~0 and~9 to common length, decompose
$C=\sum_{w\in A_k}\hat p_w^2+\sum_{w\notin A_k}\hat p_w^2$, and report the mean across seeds
of the fraction of the total change attributable to $A_k$. At $T=1$, $A_{10}$ consists of
\emph{the, to, and, of, a, in, i, that, it, he}; it accounts for 104\% of the fall in $C$,
while $A_{50}$ accounts for 108\%. Contributions above 100\% mean the remaining types offset
part of the decrease. Mean $C$ falls from $0.00784$ to $0.00410$, and the share of \emph{the}
falls from 4.94\% to 1.81\%; the total share of $A_{50}$ falls from 43.55\% to 30.88\%.
At generation~9, \emph{to} is most frequent in four seeds and \emph{of} in one; the mean of these maximum frequencies is 2.63\%.
After removing $A_{50}$ from both texts and matching the lengths of the retained streams again,
the changes in $(H_0,H_1,H_2,H_\infty)$ are $(-0.972,-0.969,-0.390,-0.686)$ bits.
This descriptive analysis selects sets by frequency and renormalizes the distribution after
deletion. Causal and grammatical effects require separate tests.

\paragraph{Prefix budget.} At $T=1$, the mean richness change and the number of seeds with
$\Delta H_0<0<\Delta H_\infty$ are shown below. The shortest tested prefixes have more types
at generation~9; in these five runs, the split occurs at every tested prefix of at least
8,000 words.
\begin{center}\small
\setlength{\tabcolsep}{3pt}
\begin{tabular}{lrrrrrrrrrr}
\toprule
words & 250 & 500 & 1,000 & 2,000 & 3,000 & 5,000 & 8,000 & 13,000 & 20,000 & 40,000 \\
$\Delta H_0$ & $+.281$ & $+.211$ & $+.195$ & $+.084$ & $-.026$ & $-.132$ & $-.258$ & $-.400$ & $-.523$ & $-.717$ \\
split seeds & 0 & 0 & 0 & 1 & 3 & 4 & 5 & 5 & 5 & 5 \\
\bottomrule
\end{tabular}
\end{center}

\paragraph{Sample coverage.} Coverage is estimated probability mass of observed types, distinct
from the fraction of population types observed in the known-law experiments. At common length
and $T=1$, estimated coverage rises from 0.907 to 0.954. Standardizing each pair to the smaller
of its two estimated coverages by coverage-based rarefaction
\citep{chao2012coverage,chao2014rarefaction} gives mean richness change $-1.91$ bits, against
$-0.81$ at equal size; it is negative in all five seeds. The mean standardized sizes are about
68,835 and 15,375 words. Coverage is estimated under exchangeable counts and continuous
interpolation of rarefaction sizes; accuracy for the underlying autoregressive process remains
untested.

\paragraph{Phrase spectra.} Common-length bigram roots for $(H_0,H_1,H_2,H_\infty)$ are
$(0.9969,0.9920,0.9784,0.9587)$, retaining the full ordering but placing $T=1$ above the split
interval. At $T=1$ their changes are $(+0.050,+0.254,+1.114,+1.631)$ bits. Trigram roots are
$(0.9975,0.9985,1.0049,1.0346)$, reversing the ordering; their $T=1$ changes are
$(+0.013,+0.015,-0.045,-0.831)$ bits. Repeated loops, including \emph{haydn haydn haydn},
concentrate the phrase head even while the unigram head disperses.

\paragraph{Simple detection baselines.} On the same 660 feature rows and 14 two-class
leave-one-setting-out folds as Table~\ref{tab:detect}, generation index alone scores mean
AUROC 0.908 and $|\Delta H_0|$ alone 0.970. Single-score orientation is chosen on the training
folds. Adding generation index to the $(H_0,H_2)$ logistic model gives 0.991. Unsigned richness scores higher than the two fixed pairs in this pool.
Appendix~\ref{app:familyholdout} gives a post-hoc family-holdout analysis; prospective transfer
to additional model families and the mechanisms behind these scores remain untested. The pooled in-sample $(H_0,H_2)$ fit has AUROC 0.909, with a
block-bootstrap interval $[0.879,0.945]$ over 48 (setting, seed) blocks.

\section{Figure values and experimental detail}
\label{app:figvals}

\paragraph{Data provenance.} English training text is five books from NLTK's Gutenberg selection
(Austen's \emph{Emma}, Carroll's \emph{Alice's Adventures in Wonderland}, Chesterton's \emph{The
Man Who Was Thursday}, Bryant's \emph{Stories to Tell to Children} and Edgeworth's \emph{The
Parent's Assistant}), with \emph{Persuasion} held out; the
$V$-bridge uses 48-token contexts from four of these books (all but \emph{The Parent's Assistant}). The four-arm study of \S\ref{sec:robust} and
the temperature sweep of Appendix~\ref{sec:tempfails} start from a common checkpoint: GPT-2 adapted
for two epochs on 700 random 256-token blocks of real text at learning rate $5\times10^{-5}$. The
temperature grids start from pretrained checkpoints.

\paragraph{Training details.} Each training corpus holds four batches of continuations (64 per
batch in the English grid), each generated from a 16-token real prefix with at most 176 new
tokens; continuations shorter than 24 tokens after padding removal are discarded. Each generation
resets AdamW and trains for two epochs on shuffled 256-token blocks in batches of eight, at
learning rate $6\times10^{-5}$ with gradient norm clipped at $1.0$, so the number of optimizer
steps varies with corpus length.

\paragraph{Non-English corpora.} \textbf{German:} Carroll \emph{Alice} (DE), Goethe \emph{Werther} and \emph{Faust~I}, Kafka \emph{Die Verwandlung}, Mann \emph{Der Zauberberg}; held out \emph{Dr.\ Mabuse}. \textbf{Spanish:} Cervantes \emph{Don Quijote}, Hugo \emph{Los miserables}, Dostoyevsky \emph{El crimen y el castigo}, Pardo Baz\'an \emph{Insolaci\'on y Morri\~na}, Unamuno \emph{Andanzas y visiones espa\~nolas} (travel essays); held out Ponson du Terrail \emph{La cuerda del ahorcado}. \textbf{Chinese:} \emph{Xiyouji}, \emph{Jin Ping Mei}, \emph{Jingshi Tongyan}, \emph{Jingu Qiguan}, \emph{Badong Tian}; held out \emph{Xu Xiake Youji} (Classical Chinese travelogue). \textbf{Hindi:} Hindi Wikisource, Premchand's \emph{Gaban}, \emph{Rangbhoomi}, \emph{Karmabhoomi}, \emph{Nirmala}, \emph{Sevasadan} and the \emph{Mansarovar} stories, with \emph{Godaan} held out. MAUVE is reported only for English, because its featurizer is trained on English.

\paragraph{Reference diversity.} Chinese character $\Htwo$ is $7.63$--$8.58$ in ten non-overlapping
200,000-character training windows and $7.57$--$7.73$ in two such held-out windows. Windows
are cut from the original Unicode text; entropy is computed on the Han characters within each. Generation-0 word values are $6.85$
(English, the adapted checkpoint of the four-arm study), $7.51$ (German) and $6.32$ (Spanish). Comparisons across languages require a common
measurement convention.

\begin{figure}[ht]
\centering
\includegraphics[width=\linewidth]{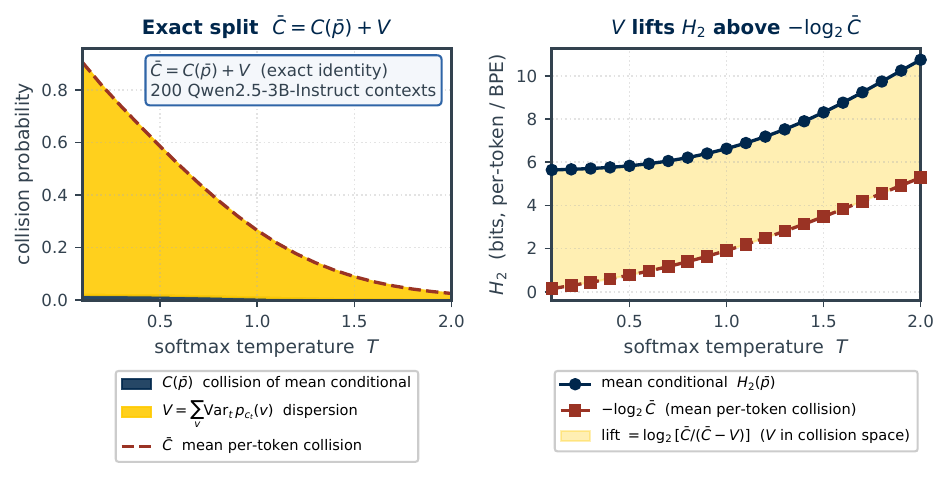}
\caption{The $V$-bridge on Qwen2.5-3B-Instruct, over 200 fixed human-text contexts and the full
$151{,}936$-token vocabulary. \emph{Left:} mean conditional collision decomposes as
$\Cbar=C(\pbar)+V$. \emph{Right:} the entropy of the mean conditional lies above $-\log_2\Cbar$.
All plotted entropies are computed from conditional softmax probabilities.}
\label{fig:vbridge}
\end{figure}

\paragraph{$V$-bridge values.} Table~\ref{tab:vbridgevals} gives the per-token BPE values for
Qwen2.5-3B-Instruct plotted in Figure~\ref{fig:vbridge}. Computing $V$ from
$\sum_v\Var_t[p_{c_t}(v)]$ agrees with $\Cbar-C(\pbar)$ to $1.1\times10^{-8}$ (limited by float32
precision), and $\Cbar$ matches separately stored per-context values to $3\times10^{-9}$.

\begin{table}[ht]
\centering\small
\caption{$V$-bridge values on 200 fixed human-text contexts (Qwen2.5-3B-Instruct, per-token BPE
distributions over the full vocabulary).}
\label{tab:vbridgevals}
\begin{tabular}{rrrrr}
\toprule
$T$ & $\Cbar$ & $V$ & $C(\pbar)$ & $\Htwo(\pbar)$\\
\midrule
0.1 & 0.90471 & 0.88476 & 0.01995 & 5.647\\
0.5 & 0.58531 & 0.56781 & 0.01749 & 5.837\\
1.0 & 0.26548 & 0.25539 & 0.01008 & 6.632\\
2.0 & 0.02557 & 0.02499 & 0.00058 & 10.751\\
\bottomrule
\end{tabular}
\end{table}

\paragraph{Compute.} Training used Apple M-series laptops with 16 or 48\,GB of
memory (Metal Performance Shaders, MPS, backend). Approximate runtimes were 21 hours for the 675 generation-steps of the coarse
grid and 22 hours for the 39 matched LoRA trajectories of the main tables (351 steps). The two refinements added 135 steps
in 4.2 hours and 90 in 2.9 hours on 48\,GB, running up to two trajectories at a time, and the
matched repetition-penalty rerun added 54 steps in 2.0 hours with three trajectories at a time.
The penalty-free runs of the other three Qwen-Instruct samplers (81 steps) and the branch
intervention (825 steps) took 15.3 hours, running four processes at a time. In total the reported
experiments before the matched-reader check took roughly 295 device-hours, excluding validation reruns.
The check added 86 minutes of inference for 120 corpora, without retraining; its manifests record
the sampling environment and hashes of every checkpoint and output text. The fresh-seed studies of
Appendix~\ref{app:freshseeds} took 22 hours of elapsed time with two concurrent workers.

\paragraph{Run provenance.} The manifests of the temperature grids, the LoRA and Qwen-Instruct
settings and the Pythia study record pinned model revisions, corpus hashes, library versions
and adapter fingerprints. For eleven inventoried main settings and the count-only controls,
frozen-detector, EOS and corpus-size runs, the revision pins in the release were inferred after the runs; the original
revisions were not recorded in manifests and remain uncertain. Spanish GPT-2 was
loaded from its floating main branch, whose head has been commit \texttt{69e8e9b} since April
2024; the release pins that commit. The exact Qwen2.5-3B-Instruct revision used for the
$V$-bridge remains unrecorded. A download log records commit \texttt{aa8e725} of Qwen2.5-3B-Instruct for an
earlier next-token experiment.

The four-arm and temperature-sweep trainers saved checkpoints without generation markers.
An extra update during a mid-trajectory resume cannot be ruled out from the retained records.
MPS nondeterminism further limits exact reproduction.

\clearpage
\section{Fresh-seed studies and family-held-out analysis}
\label{app:freshseeds}

\subsection{Prospective replication of the primary crossing order}

We fixed the eleven-temperature grid of \S\ref{sec:crossings}, five fresh training seeds
(100--104), both text budgets and the analysis before acquisition. The 55 trajectories
start from pretrained GPT-2 and use nine full fine-tuning generations at training top-$p=1$,
with the primary model revision, corpus and training settings. These include two epochs
of AdamW at $6\times10^{-5}$, batch size eight, an optimizer reset each generation and
gradient clipping at norm one. Measurements use $T=1$, top-$p=0.95$, top-$k=0$ and repetition
penalty one, with eight batches of 64 continuations. The frozen protocol records the model
and tokenizer revision, all sampler settings and corpus hashes.

For each seed and temperature, we compare generation~9 with generation~0 at their common
word count and at the first 13,000 words. Tokenization is unchanged. The first upward root
is linearly interpolated between adjacent measured temperatures; intervals resample whole
training seeds jointly across temperatures, using 4,000 draws. The point roots and intervals
are in Table~\ref{tab:freshcrossings}. All four roots exist in every draw at both budgets, and
$T^*(\Hinf)<T^*(\Htwo)<T^*(H_1)<T^*(\Hz)$ holds in 4,000/4,000 draws under each budget.
Every seed also has $\Delta\Hz<0<\Delta\Hinf$ at $T=1$ under both budgets.

\begin{table}[ht]
\centering\small
\caption{Fresh-seed replication: first upward crossing temperatures and 95\% paired-seed
bootstrap intervals. The same five resampled seed indices are used across all temperatures.
Intervals condition on the selected grid and exclude interpolation error.}
\label{tab:freshcrossings}
\begin{tabular}{lrrrr}
\toprule
& \multicolumn{2}{c}{Common endpoint length} & \multicolumn{2}{c}{13,000 words} \\
Order & root & 95\% interval & root & 95\% interval \\
\midrule
$\Hinf$ & $0.906$ & $[0.871,0.937]$ & $0.903$ & $[0.856,0.944]$ \\
$\Htwo$ & $0.962$ & $[0.955,0.969]$ & $0.963$ & $[0.951,0.971]$ \\
$H_1$ & $0.999$ & $[0.996,1.002]$ & $0.993$ & $[0.990,0.996]$ \\
$\Hz$ & $1.032$ & $[1.026,1.038]$ & $1.020$ & $[1.016,1.026]$ \\
\bottomrule
\end{tabular}
\end{table}

Across the 18 sampled orders listed in Appendix~\ref{app:posthocanalyses}, each measured
temperature has changes of one sign or a single negative-to-positive transition, in every seed
and in the mean, at both budgets. Each seed's interpolated roots decrease strictly across
these sampled orders. This replication adds independent training seeds within the already
selected model, corpus and temperature grid; transfer across settings is addressed by the
separate experiments in the main text.

\subsection{Matched controls under the primary measurement protocol}
\label{app:freshcontrols}

We reuse the five new $T=1$ recursive trajectories and add five human-text adaptations and
five unchanged-model controls with the same seed indices. At each generation, a human
adaptation draws fresh blocks with replacement from the original five-book training pool.
Its paired recursive run determines the number of 256-token blocks and optimizer updates;
learning rate, batch size, epochs, optimizer reset and gradient clipping are identical.
Human block sampling has a separate recorded random-number stream. All three arms share
the prompt pool and keyed measurement streams, and all entropies use the first 13,000 words.
The frozen arm changes only the measurement seed across generations and takes no optimizer steps.

We retain the one-bit magnitude rule, the profile rule with its $0.20$-bit sign floor and the
unfloored rule from \S\ref{sec:detect}. A separately specified $0.02$-bit floor tests sensitivity
to smaller sign disagreements. The thresholds were fixed before data collection. A trajectory
is above the quality cutoff if any generation~1--9 reaches twice its generation-0 perplexity
on \emph{Persuasion}. All five recursive runs cross that cutoff; all ten human and frozen
controls stay below it. Table~\ref{tab:freshcontrols} gives every arm's alarm counts.

\begin{table}[ht]
\centering\small
\caption{Trajectories ever flagged in generations~1--9 under the primary reader, five seeds
per arm. Recursive runs cross the reference-likelihood cutoff; human and frozen runs stay below
it. All 150 measurement texts reach the 13,000-word budget.}
\label{tab:freshcontrols}
\begin{tabular}{lrrr}
\toprule
Rule & Recursive & Human adaptation & Frozen \\
\midrule
Magnitude only, $M_g>1$ bit & $3/5$ & $0/5$ & $0/5$ \\
Profile, $0.20$-bit sign floor & $5/5$ & $5/5$ & $0/5$ \\
Profile, $0.02$-bit sign floor & $5/5$ & $5/5$ & $5/5$ \\
Profile without a sign floor & $5/5$ & $5/5$ & $5/5$ \\
\bottomrule
\end{tabular}
\end{table}

The $0.20$-bit floor filters all five frozen controls, yet flags all five human adaptations
below the likelihood cutoff. Magnitude alone avoids these ten false alarms but misses two of
the five degraded recursive runs. Thus the ambiguity seen under the secondary sampler also
occurs under the primary protocol with matched update budgets. These are counts for five
seeds per arm and a stated reference criterion, rather than deployment false-positive rates.

The released records include every generation's perplexity on \emph{Persuasion},
\emph{Sense and Sensibility}, \emph{Moby Dick} and \emph{The Adventures of Buster Bear}.
The primary work uses its original first 60 blocks; each additional work uses 60 equally
spaced disjoint blocks of the same length. All four use the primary MPS loss helper, with
token hashes and block positions recorded. Training uses none of these held-out works.
Each of the three additional works gives the same trajectory labels: all five recursive
runs cross its twice-baseline cutoff, and all ten controls stay below it.

\subsection{Holding out complete model families}
\label{app:familyholdout}

This separate post-hoc analysis groups the existing 660-row detection pool by model family
(Table~\ref{tab:familyholdout}).
The GPT-2 fold contains 381 rows (178 above the cutoff, 203 below), Qwen2.5 has 171
(81 above, 90 below), and SmolLM2 has 108 (18 above, 90 below). Each outer fold withholds
every setting, seed, decoder and generation of one family. On its two remaining families,
nested selection considers all 15 feature pairs and maximizes mean leave-one-family-out AUROC,
breaking exact ties lexicographically. Standardization is fitted only to training rows.
Logistic regression uses L2 regularization, $C=1$ and the same fitting procedure as the
setting-level analysis. The signs of the absolute-richness and generation-index scores are
chosen from outer-training AUROC alone.

\begin{table}[ht]
\centering\small
\setlength{\tabcolsep}{4pt}
\caption{AUROC when a whole model family is withheld. The last row weights the three families equally and averages unrounded values. All predeclared methods are shown.}
\label{tab:familyholdout}
\begin{tabular}{lrrrrr}
\toprule
Held-out family & Nested & $\Hz+\Htwo$ & $\Htwo+\text{distinct-}2$ & $|\Delta\Hz|$ & Generation \\
\midrule
GPT-2 & $0.871$ & $0.852$ & $0.840$ & $0.888$ & $0.778$ \\
Qwen2.5 & $0.936$ & $0.941$ & $0.895$ & $0.838$ & $0.885$ \\
SmolLM2 & $0.999$ & $0.999$ & $1.000$ & $0.999$ & $0.900$ \\
\midrule
Equal-family mean & $0.935$ & $0.931$ & $0.912$ & $0.908$ & $0.855$ \\
\bottomrule
\end{tabular}
\end{table}

Nested selection chooses $(\Hz,\Hinf)$ when GPT-2 is withheld,
$(\Hinf,\text{distinct-}2)$ for Qwen2.5 and $(\Hz,H_1)$ for SmolLM2.
The equal-family means average different test populations from the 14 scoreable setting folds
in \S\ref{sec:detect}. They are a sensitivity analysis on three already-inspected families.
The release includes all out-of-fold predictions, per-setting scores with explicit missing
values for single-class settings, inner-selection traces and fitted parameters.

\paragraph{Compute and evidence.} The studies share five $T=1$ trajectories and total 65
trajectories, 540 trained generation steps and 650 texts, using at most two MPS workers
on 48\,GB. Generation markers and atomic checkpoints support resumption. The release's
\texttt{code/exp\_data/fresh\_seed\_study/} directory contains the frozen protocols, code, all measurements,
unigram counts and text hashes, plus the 250 analysis texts deduplicated by hash. The other
400 intermediate grid texts and all 60 trained endpoint checkpoints remain retained locally;
checkpoint digests are released. A CPU verifier reconstructs the tables, profiles and
bootstraps. Likelihood inference and retraining require model weights.

\end{document}